\documentclass[letterpaper]{article}
\usepackage[preprint]{kosmo}
\usepackage[hyphens]{url} 
\usepackage{graphicx}
\def\UrlFont{\rm}
\usepackage{natbib} 
\usepackage{caption}
\usepackage{booktabs}
\usepackage{array}
\usepackage{amsmath,amssymb,amsfonts,mathtools}
\usepackage{placeins}
\usepackage[most]{tcolorbox}
\usepackage{xcolor}
\usepackage{environ}
\definecolor{kosmoblue}{HTML}{315FD5}
\definecolor{kosmofg}{HTML}{17191D}
\definecolor{kosmobg}{HTML}{F1F4F7}
\definecolor{kosmomuted}{HTML}{4B5159}
\definecolor{kosmoline}{HTML}{17191D}
\newcommand{\projectpageurl}{{\def\UrlFont{\ttfamily}\url{https://kosmoresearch.github.io/UniQuery4R/}}}

\title{UniQuery4R: Unified 4D Scene Reconstruction from a Single Query}
\author{
    Tiancheng Chen\textsuperscript{\rm 1},
    Sheng Tang\textsuperscript{\rm 1},
    Wenhua Jin\textsuperscript{\rm 1,2},
    Weiqi Zhang\textsuperscript{\rm 3},
    Juntong Fang\textsuperscript{\rm 3},\\
    Junsheng Zhou\textsuperscript{\rm 3},
    Zesong Li\textsuperscript{\rm 1}
}
\affiliations{
    \textsuperscript{\rm 1}Kosmo Research\quad
    \textsuperscript{\rm 2}Automotive Engineering Department, Jilin University\quad
    \textsuperscript{\rm 3}School of Software, Tsinghua University
}

\makeatletter
\RenewEnviron{abstract}{\global\let\@kosmo@abstract\BODY}
\renewcommand{\maketitle}{%
  \par
  \begingroup
    \def\thefootnote{\fnsymbol{footnote}}
    \twocolumn[{%
      \vspace*{-0.30in}%
      \noindent
      \includegraphics[width=2.75cm]{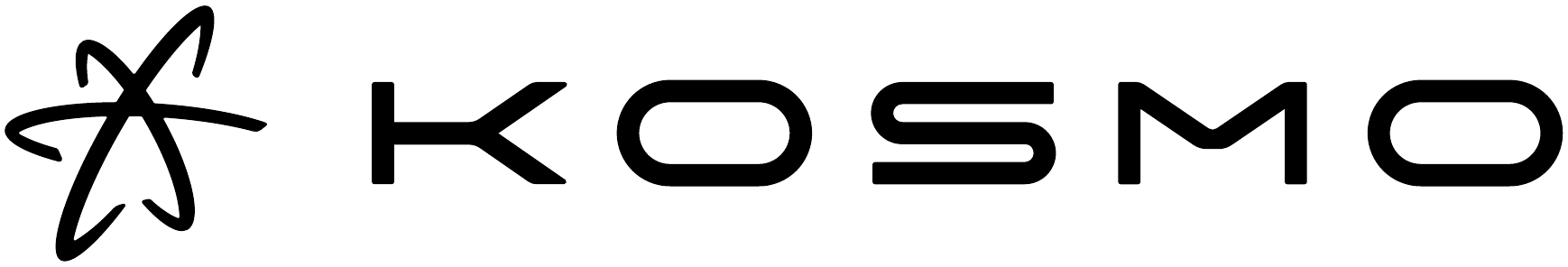}\par
      \vspace{0.15cm}
      {\color{kosmoline}\hrule height 0.45pt}
      \vspace{0.30cm}
      \centering
      {\fontsize{20}{24}\selectfont\sffamily\bfseries
        \color{kosmofg}\@title\par}
      \vspace{0.38cm}
      {\fontsize{10.8}{13}\selectfont\sffamily\mdseries
        \color{kosmofg}\@author\par}
      \vspace{0.15cm}
      {\small\color{kosmomuted}\kosmo@affiliations\par}
      \vspace{0.32cm}
      \begin{tcolorbox}[
        enhanced,
        frame hidden,
        colback=kosmobg,
        coltext=kosmofg,
        arc=8pt,
        left=0.60cm,
        right=0.60cm,
        top=0.42cm,
        bottom=0.40cm,
        boxsep=0pt,
        before skip=0pt,
        after skip=0.48cm,
        grow to left by=1.5pt,
        grow to right by=1.5pt
      ]
        \raggedright
        {\normalsize\sffamily\bfseries\color{kosmofg}Abstract\par}
        \vspace{0.2em}
        {\small\renewcommand{\baselinestretch}{1.04}\selectfont
          \color{kosmofg}\@kosmo@abstract\par}
        \vspace{0.35cm}
        {\small\sffamily\bfseries\color{kosmofg}Project page:\ }%
        {\small\color{kosmoblue}\projectpageurl\par}
      \end{tcolorbox}
    }]%
    \long\def\@footnotetext##1{\insert\kosmo@thanksins{%
        \protect\footnotesize\interlinepenalty\interfootnotelinepenalty
        \splittopskip\footnotesep\splitmaxdepth\dp\strutbox
        \floatingpenalty\@MM\hsize\columnwidth\@parboxrestore
        \protected@edef\@currentlabel{%
           \csname p@footnote\endcsname\@thefnmark}%
        \color@begingroup
          \@makefntext{%
            \rule\z@\footnotesep\ignorespaces##1\@finalstrut\strutbox}%
        \color@endgroup}}%
    \@thanks
  \endgroup
  \if T\copyright@on\insert\kosmo@copyrightins{\noindent\footnotesize\copyright@text}\fi
  \setcounter{footnote}{0}%
  \let\maketitle\relax
  \let\@maketitle\relax
  \gdef\@thanks{}%
  \gdef\@author{}%
  \gdef\@title{}%
  \let\thanks\relax
}
\makeatother

\begin{document}

\begin{abstract}
Reconstructing dynamic 4D scenes requires jointly estimating correspondence, geometry, object motion, and camera motion. Existing feed-forward methods typically predict dense task-specific maps or independently process source--target pairs, leading to unnecessary computation for sparse queries and limited feature reuse across different frame pairs. We present \textbf{UniQuery4R}, a query-conditioned framework that encodes a multi-frame clip once and selects the source view, target view, and continuous source-image coordinate only at decoding time via source-to-target cross-attention. Each query jointly predicts target correspondence, target-time 3D position, and scene flow, along with source depth, while camera parameters are estimated per view. This design allows the encoded clip to be reused across arbitrary source--target selections and supports both sparse inference and dense reconstruction through batched queries, without learned temporal embeddings tied to a fixed clip length. We further introduce a direction--magnitude parameterization of scene flow with separate supervision for moving and static points. Among the evaluated methods, UniQuery4R achieves the best macro-average results on WorldTrack for both scene-flow estimation and dynamic-point reconstruction.
\end{abstract}

\maketitle

\section{Introduction}
\label{sec:intro_4d_feedforward}

Estimating scene geometry, object motion, and camera motion from video is a fundamental problem in dynamic scene understanding. Classical SfM, SLAM, and MVS pipelines are effective for static scenes~\cite{schoenberger2016colmap,yao2018mvsnet}, but their reliance on scene rigidity and iterative optimization limits their applicability to unconstrained dynamic videos.

Recent feed-forward methods predict different combinations of dynamic geometry, object motion, and camera pose~\cite{lin2025movies,yang2025neoverse}. Many of them produce dense outputs or construct global scene representations, which are well suited to full-scene reconstruction but incur unnecessary computation when only a small set of points is queried. Moreover, correspondence, geometry, and motion are often handled by task-specific or sequential modules rather than a shared point-level representation.

We instead formulate dynamic reconstruction around a continuous source-pixel query. Tracking datasets annotate 2D trajectories with floating-point image coordinates; dense per-pixel prediction on an integer grid quantizes these labels and loses sub-pixel information, whereas a continuous query reads the requested location directly. Given a source-image coordinate and a target view, the model predicts the point's target correspondence, target-time 3D position, and scene flow from a shared query representation. These quantities describe the same physical point across space and time and can therefore be estimated from common visual evidence. Camera parameters, by contrast, are view-level quantities and are predicted separately from per-frame representations.

% Page-1 top-right placement: declare after left-column intro text so [t] floats to the right column.
\begin{figure}[t]
\centering
\includegraphics[width=0.9\columnwidth,keepaspectratio]{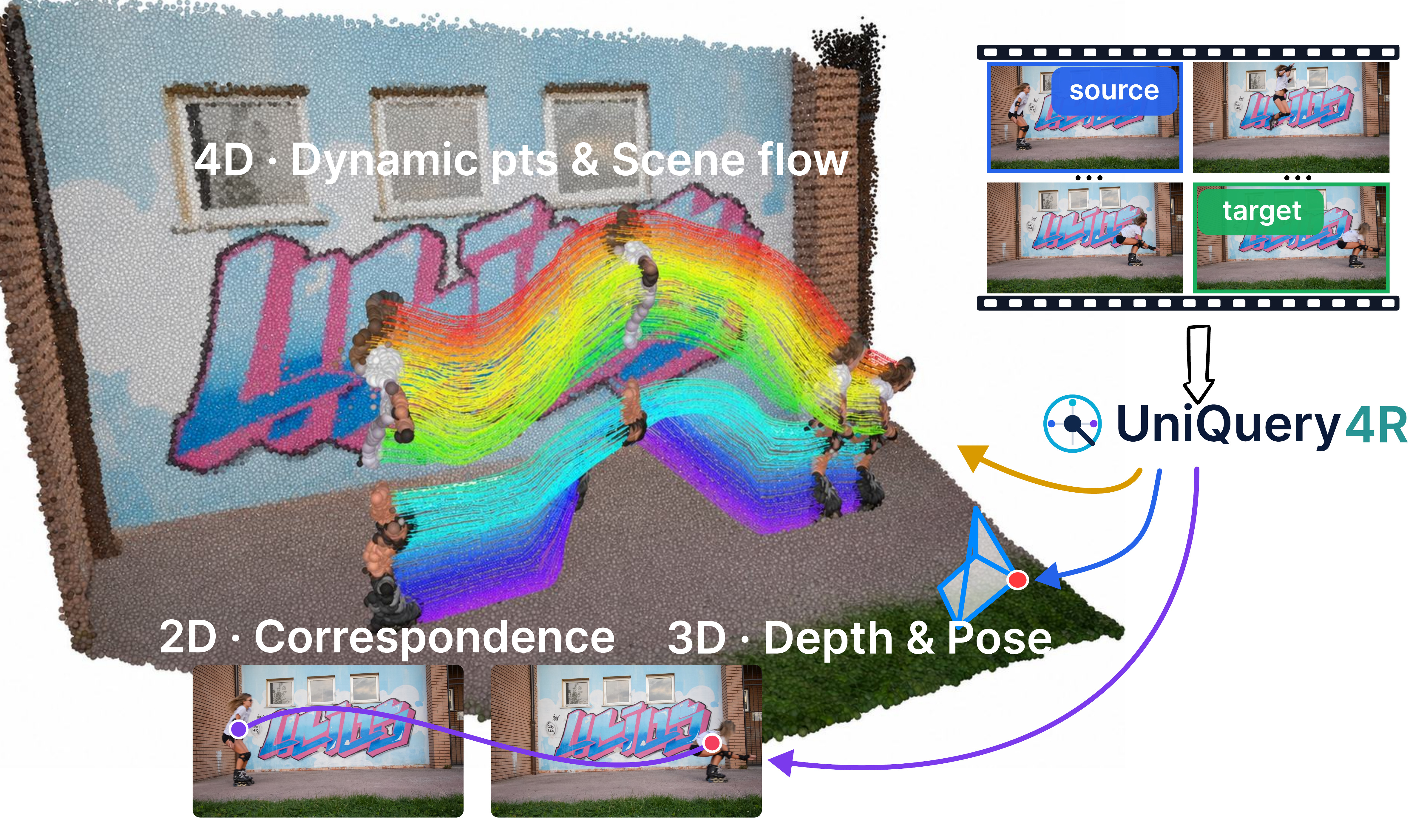}
\caption{Conceptual overview. A continuous source-pixel query over a jointly encoded clip predicts target correspondence, target-time 3D position, scene flow, and source depth, while camera parameters are estimated per view.}
\label{fig:concept}
\end{figure}

We introduce \textbf{UniQuery4R}, which jointly encodes the input clip independently of the query $\mathbf{q}=(u,v,s,t)$; the source view, target view, and source-image coordinate are used only by the query decoder (Figure~\ref{fig:concept}). In D4RT~\cite{zhang2025d4rt} (Table~\ref{tab:paradigm_comparison}), each query attends to a global scene representation comprising patches from all input frames. Because this interaction alone does not identify the requested target frame, D4RT specifies the target time through a learned temporal embedding in the query, tying the model to a predefined 48-frame horizon. UniQuery4R instead samples a query token at the specified coordinate from the selected source features and lets it attend only to the selected target features. The target frame is thus specified by feature selection rather than a learned temporal code, avoiding fixed-index target-time embeddings and enabling variable-length clips.

\begin{table*}[t]
\centering
\small
\begin{tabular*}{\textwidth}{@{\extracolsep{\fill}}lcccc@{}}
\toprule
Method
& Input Clip Length
& Query Source
& Key / Value
& Supported Output \\
\midrule
V-DPM
& Variable
& N/A
& N/A
& Dense \\
D4RT
& Fixed (48 frames)
& Input embedding
& All-frame patches
& Sparse / Dense \\
\textbf{UniQuery4R (ours)}
& \textbf{Variable}
& \textbf{Sampled from $F_s$}
& \textbf{$F_t$ patches}
& \textbf{Sparse / Dense} \\
\bottomrule
\end{tabular*}
\caption{Paradigm comparison with D4RT and V-DPM. UniQuery4R differs in
temporal horizon, query formation, and target-side attention. ``Variable''
means no learned fixed clip length (memory-bounded in practice).}
\label{tab:paradigm_comparison}
\end{table*}

Our contributions are summarized as follows:
\begin{itemize}
\item We formulate feed-forward 4D reconstruction as continuous source-pixel queries over a jointly encoded, variable-length clip. Each query is formed from the selected source features and interacts only with the selected target features, avoiding temporal embeddings tied to a fixed set of frame indices. The resulting point-level representation jointly predicts correspondence, geometry, and motion, while source depth and per-view camera parameters are decoded separately.
\item We introduce a direction--magnitude parameterization of scene flow, consisting of an $\epsilon$-normalized direction vector and a non-negative magnitude. It is trained with objectives for log-space displacement, moving-point direction, and static-point magnitude, and empirically outperforms direct Cartesian regression in our ablations.
\item We evaluate UniQuery4R on four datasets under the WorldTrack protocol. Among the evaluated methods, it achieves the best four-dataset macro-average results for both scene flow and dynamic-point tracking.
\end{itemize}

\section{Related Work}

\subsection{Feed-Forward 3D and 4D Reconstruction}
\label{sec:related_3d_recon}

Classical 3D reconstruction is grounded in multi-view geometry. SfM and SLAM estimate camera motion together with scene structure, whereas MVS recovers dense geometry from calibrated views~\cite{schoenberger2016colmap,yao2018mvsnet}. Learned feature matching and differentiable optimization have further improved the robustness and accuracy of these pipelines~\cite{sarlin2020superglue,lindenberger2023lightglue,teed2021droidslam}. Despite these advances, most classical reconstruction pipelines assume a static scene and retain iterative optimization, limiting their applicability to dynamic videos.

Feed-forward models instead infer geometry and camera parameters directly from images. DUSt3R showed that dense pointmaps can be recovered from unposed pairs without SfM~\cite{wang2024dust3r}, and MUSt3R/MASt3R-SfM extended this to multi-view and unconstrained settings~\cite{cabon2025must3r,duisterhof2025mast3rsfm}; VGGT unified camera, depth, point-map, and point-track prediction in one pass~\cite{wang2025vggt}, later scaled by VGGT-$\Omega$~\cite{wang2026vggtomega} and extended to online SLAM~\cite{maggio2025vggtslam}, while MonST3R generalized pointmaps to dynamic videos~\cite{zhang2024monst3r}.

For dynamic scenes, Dynamic Point Maps (DPM) and V-DPM introduced temporally consistent pointmaps that jointly model static and dynamic geometry over time~\cite{sucar2025dpm,sucar2026vdpm}, and D2USt3R folded temporal correspondence into 4D pointmaps~\cite{han2025d2ust3r}. Recent feed-forward frameworks recover different combinations of geometry, motion, correspondence, and appearance for dynamic reconstruction~\cite{karhade2025any4d,lin2025movies,fang2025more,yang2025neoverse,zhang2025d4rt,luo20264rc,jiang2026omnix}; related foundation models such as Dens3R focus instead on static 3D geometry prediction~\cite{fang2026dens3r}. Most of these systems construct dense full-scene representations. Two conditional designs are closest to ours. D4RT answers point-level spatiotemporal queries that specify source time, target time, and camera reference, but identifies the target through a learned temporal embedding over a fixed frame horizon~\cite{zhang2025d4rt}. 4RC exposes a conditional interface that selects a target time from a reconstructed 4D representation and decodes dense geometry and motion~\cite{luo20264rc}. In contrast, we jointly encode a variable-length clip, select source and target only at decoding time, and answer each continuous pixel query $(u,v)$ by source-to-target cross-attention, with per-frame cameras decoded separately as in VGGT~\cite{wang2025vggt}. Related continuous decoding appears in InfiniDepth, which also uses a feature-pyramid decoder, but for single-image depth rather than multi-frame source-to-target queries~\cite{yu2026infinidepth}.

\subsection{3D Point Tracking}
\label{sec:related_3d_tracking}

Tracking establishes consistent cross-frame correspondences and underpins dynamic scene understanding. Recent feed-forward methods differ mainly in how they couple trajectories with geometry. SpatialTrackerV2 jointly estimates depth and camera motion, then refines tracks and poses by differentiable joint optimization that decomposes world motion into geometry, ego-motion, and object motion~\cite{xiao2025spatialtrackerv2}. St4RTrack predicts paired pointmaps in a shared world frame and chains anchored pairs for long-range correspondence~\cite{feng2025st4rtrack}. Trace Anything forms a dense trajectory field by predicting per-pixel B-spline control points for continuous-time 3D trajectories in one pass~\cite{liu2025traceanything}. Track4World globally encodes the video and estimates dense pairwise 2D/3D flow to obtain world-centric all-pixel trajectories~\cite{lu2026track4world}. Reconstruction-oriented variants include TrajVG, which couples sparse camera-frame 3D trajectories with local pointmaps~\cite{miao2026trajvg}, and Uni4D, which combines pretrained depth, tracking, and segmentation in multi-stage optimization~\cite{yao2025uni4d}. These works emphasize full trajectories or scene-level reconstruction; we instead answer a continuous source-pixel query at a chosen target by jointly decoding correspondence, target-time 3D geometry, and scene flow.

\subsection{Scene Flow and Dense 4D Motion}
\label{sec:related_scene_flow}

Scene flow is the 3D displacement of scene points between observations and links dynamic geometry to tracking. MoVieS models time-varying motion of pixel-aligned Gaussians for view synthesis, geometry, and zero-shot scene flow~\cite{lin2025movies}; Any4D predicts dense metric per-pixel geometry and motion over multiple frames~\cite{karhade2025any4d}; and 4RC encodes once then conditionally decodes dense geometry and motion to arbitrary target times~\cite{luo20264rc}. Track4World estimates dense pairwise 2D/3D flow from a globally encoded video~\cite{lu2026track4world}. Other representations encode motion differently: V-DPM predicts time-variant and time-synchronized dynamic pointmaps from which per-point 3D motion can be recovered~\cite{sucar2026vdpm}, while OmniX parameterizes dense trajectory fields with compact dynamic tokens that separate dynamic from static geometry~\cite{jiang2026omnix}. Flow4R treats camera-space scene flow as the central quantity, predicting per-pixel point, flow, pose weight, and confidence from each image pair and processing sequences as independent anchor--frame pairs~\cite{qian2026flow4r}, so each pair needs a new two-view pass rather than reusing one jointly encoded multi-view clip.

\section{Method}
\label{sec:method}

\subsection{Overview}
\label{sec:overview}

\begin{figure*}[t]
  \centering
  \includegraphics[width=\textwidth,keepaspectratio]{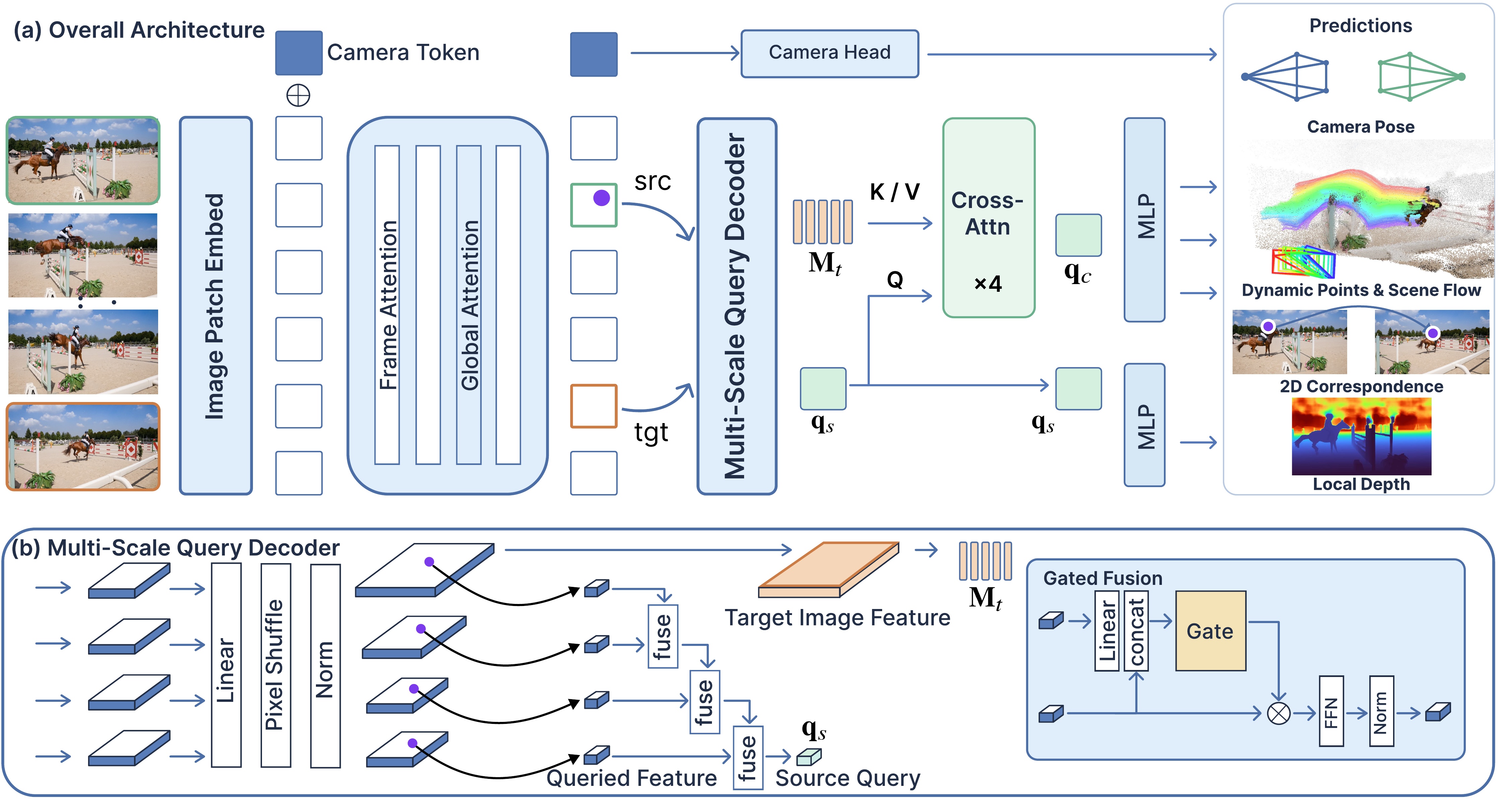}
  \caption{(a)~Overall architecture. A clip is patch-embedded with camera tokens and jointly encoded by frame/global attention. At decoding time, $\mathbf q=(u,v,s,t)$ selects source/target; the multi-scale query decoder yields $\mathbf{q}_s$ and target features $\mathbf{M}_t$, then $\mathbf{q}_s$ cross-attends to $\mathbf{M}_t$ ($\times4$) to form $\mathbf{q}_c$. Src$\to$Tgt MLPs predict $\mathbf{P}$, $\Delta\mathbf{P}$, and $\mathbf{f}$ from $\mathbf{q}_c$; a Src-only MLP predicts local depth from $\mathbf{q}_s$; a camera head reads per-view tokens. (b)~Multi-scale query decoder with gated fusion that builds $\mathbf{q}_s$ from the source pyramid.}
  \label{fig:model_arch}
\end{figure*}

Given a clip $\mathcal I=\{I_n\}_{n=0}^{N-1}$, UniQuery4R first jointly encodes all $N$ views into multi-view-contextualized feature pyramids. A continuous query $\mathbf q=(u,v,s,t)$ selects source $I_s$ and target $I_t$ only at decoding time. Multi-scale sampling from the selected source yields $\mathbf{q}_s$, which attends to all tokens of the selected target to form $\mathbf{q}_c$ for correspondence, geometry, and motion; source features provide depth, while per-frame tokens predict cameras (Figure~\ref{fig:model_arch}). Source and target views are selected from the encoded features only at decoding time, allowing the backbone computation to be reused across different $(s,t)$ choices. The query decoder represents temporal relations through source-to-target feature interaction rather than a learned temporal embedding indexed by a predefined frame set.

The query interface preserves sub-pixel coordinates through differentiable sampling. At a fixed target resolution, decoding cost scales with the requested number of points, so sparse on-demand inference and dense reconstruction share the same model, with the latter obtained by batching a grid of queries. Because one query-conditioned representation drives correspondence, geometry, and motion, these outputs can exploit common evidence instead of being reconstructed by isolated task pipelines.

\subsection{Query Formulation and Outputs}
\label{sec:problem_formulation}

\begin{table}[t]
\centering
\small
\begin{tabular}{@{}l@{\hskip5pt}l@{\hskip5pt}c@{\hskip5pt}l@{}}
\toprule
Output & Sym. & Q? & Description \\
\midrule
\multicolumn{4}{@{}l}{\textit{Local, per-query --- from $\mathbf{q}_c$}}\\
Dynamic point & $\mathbf{P}\!\in\!\mathbb{R}^3$       & Y & target-time pos.\ (ref.-cam.) \\
Scene flow    & $\Delta\mathbf{P}\!\in\!\mathbb{R}^3$ & Y & $s\!\to\!t$ displ.\ (ref.-cam.) \\
2D corresp.   & $\mathbf{f}\!\in\!\mathbb{R}^2$       & Y & target-image coordinate \\
Conf./prec.   & $(c_P,\mathbf{c}_f)$                   & Y & point and correspondence \\
\midrule
\multicolumn{4}{@{}l}{\textit{Local, per-query --- from $\mathbf{q}_s$}}\\
Depth/conf.   & $(d,c_d)$                              & Y & source depth and confidence \\
\midrule
\multicolumn{4}{@{}l}{\textit{Global, per-view --- camera tokens}}\\
Camera        & $\boldsymbol{\pi}\!\in\!\mathbb{R}^9$  & N & extrinsics + intrinsics \\
\bottomrule
\end{tabular}
\caption{UniQuery4R outputs for query $\mathbf{q}=(u,v,s,t)$. ``Q?'' denotes query dependence; camera parameters are predicted per view.}
\label{tab:outputs}
\end{table}

A query $\mathbf{q}=(u,v,s,t)$ specifies a pixel $(u,v)$ in source view $I_s$ and a target view $I_t$. UniQuery4R predicts
\begin{equation}
Q(u,v,s,t)=\{\mathbf{P},\Delta\mathbf{P},\mathbf{f},c_P,\mathbf{c}_f,d,c_d\},
\label{eq:query_outputs}
\end{equation}
together with query-independent per-view camera parameters $\boldsymbol{\pi}$. We fix view~$0$, rather than the source view, as the reference. In this frame, $\mathbf{P}\in\mathbb{R}^3$ is the target-time location of the queried source point and $\Delta\mathbf{P}\in\mathbb{R}^3$ is its $s\!\to\!t$ scene flow. Pixel coordinates are mapped to $[-1,1]^2$ for sampling. The correspondence $\mathbf{f}\in\mathbb{R}^2$ is directly regressed as the absolute target coordinate in this normalized system; its linear output is not hard-clipped to $[-1,1]^2$. The supervised auxiliary outputs are point confidence $c_P$, four correspondence confidence/precision logits $\mathbf{c}_f\in\mathbb{R}^4$, and depth confidence $c_d$ (Table~\ref{tab:outputs}). Finally, $\boldsymbol{\pi}$ contains per-view translation, quaternion rotation, and FoVs ordered as vertical then horizontal $(\mathrm{FoV}_v,\mathrm{FoV}_h)$.

\paragraph{Coordinate and Scale Convention.}
All 3D outputs and camera translations share the view-$0$ coordinate system and scale. During training, valid ground-truth geometry is divided by the clip-level mean point-to-view-$0$-origin distance. At inference this label-derived scale is unavailable, so predictions are up to a global scale. Following WorldTrack, point maps and scene flow are independently aligned using the same median-based procedure: for each output type, the global scale is the ratio of the median ground-truth Euclidean magnitude to the median predicted Euclidean magnitude. The $0.02$\,m moving/static threshold is applied after restoring each sample's metric scale, whereas displacement regression operates in the normalized space.

\subsection{Multi-Scale Source-to-Target Attention}
\label{sec:query_crossattn}

UniQuery4R uses an internal multi-view 3D foundation model initialized from DA3-Giant~\cite{lin2025depthanything3}. It jointly processes $\mathcal I$ so that each view's tokens are contextualized by the other input views. Tokens from four encoder stages are linearly projected and pixel-shuffled with stage-dependent upsampling factors $s_\ell$:
\begin{equation}
\mathbf{F}_n^{(\ell)}
=
\mathrm{PixelShuffle}_{s_\ell}\!\left(\mathrm{Linear}(\mathbf{X}_n^{(\ell)})\right),
\qquad \ell=1,\ldots,4.
\label{eq:pixel_shuffle}
\end{equation}
The resulting multi-scale feature pyramid $\{\mathbf{F}_n^{(\ell)}\}$ combines fine localization from shallow stages with semantic and multi-view context from deeper stages. Source and target pyramids are selected from this shared representation only after encoding; further encoder details are provided in the supplementary material.

Given a query $(u,v)$, we bilinearly sample each level of the source pyramid at the continuous location, preserving sub-pixel coordinates without quantization. We order the sampled features from the shallowest/finest to the deepest as $\{\mathbf{f}^{(k)}\}_{k=1}^{4}$. Starting from the finest sampled feature, we progressively inject deeper semantics through learned gates:
\begin{equation}
\begin{aligned}
&\mathbf{h}^{(1)}
=
\mathbf{f}^{(1)},\\
&\widetilde{\mathbf{h}}^{(k-1)}
=
\mathbf{W}_h^{(k)}\mathbf{h}^{(k-1)},\\
&\mathbf{g}^{(k)}
=
\sigma\!\left(
\mathbf{W}_g^{(k)}
[\widetilde{\mathbf{h}}^{(k-1)};\mathbf{f}^{(k)}]
\right),\\
&\mathbf{z}^{(k)}
=
\mathbf{g}^{(k)}\odot\widetilde{\mathbf{h}}^{(k-1)}
+(1-\mathbf{g}^{(k)})\odot\mathbf{f}^{(k)},\\
&\mathbf{h}^{(k)}
=
\mathrm{FFN}^{(k)}\!\left(\mathrm{LN}^{(k)}(\mathbf{z}^{(k)})\right),
\quad k=2,\ldots,4.
\end{aligned}
\label{eq:gated_fusion}
\end{equation}
Here $\mathbf{W}_h^{(k)}$ projects the accumulated feature to the current level, and the gate adaptively retains fine localization while incorporating progressively deeper semantic and multi-view context. The fused feature $\mathbf{q}_s=\mathbf{h}^{(4)}$ is the continuous source query representation.

For source--target interaction, the query indices $(s,t)$ select two feature hierarchies from the already encoded clip, and $\mathbf{q}_s$ attends to the entire target feature map $\mathbf{M}_t=\mathrm{Flatten}(\mathbf{F}_t^{(4)})$. We apply a stack of Pre-LN cross-attention blocks in which $\mathbf{q}_s$ is the query and $\mathbf{M}_t$ provides keys and values, followed by a linear readout to the query-conditioned representation $\mathbf{q}_c$. Crucially, we do not sample the target at the same $(u,v)$: correspondence emerges from attending over the full target field. The module uses neither local windows, predefined matches, nor a temporal embedding. Thus, $(s,t)$ is a decoding choice rather than an input-pair choice, and different source--target relations can be queried from the same jointly encoded clip.

\subsection{Geometry and Correspondence Heads}
\label{sec:geometry_motion}

Shallow task-specific MLPs decode $\mathbf{P}$, $\Delta\mathbf{P}$, $\mathbf{f}$, point confidence, and correspondence confidence/precision from $\mathbf{q}_c$, while $\mathbf{q}_s$ produces depth and its confidence. We use an inverse-log parameterization for $\mathbf{P}$ and an exponential map for depth; $\mathbf{f}$ is a linear absolute target coordinate $\mathbf{f}^*\in[-1,1]^2$ rather than a displacement from the source query. Separate supervision with shared query representations encourages the tasks to exploit common spatiotemporal evidence. Cameras are predicted per view by a VGGT-style token head~\cite{wang2025vggt} and supervised with reference-normalized relative $L_1$ on translation, quaternion (canonicalized to $w\ge0$), and FoV, following the relative-pose supervision principle of~$\pi^3$~\cite{wang2026pi3}.

\subsection{Direction--Magnitude Scene Flow and Supervision}
\label{sec:scene_flow_modeling}

Rather than directly regressing the three Cartesian components of scene flow, we factorize the displacement into an $\epsilon$-normalized direction vector and a non-negative magnitude:
\begin{equation}
\begin{aligned}
m&=\mathrm{softplus}(\tilde{m}),\qquad
\hat{\mathbf{v}}=
\frac{\tilde{\mathbf{v}}}
{\sqrt{\|\tilde{\mathbf{v}}\|_2^2+\epsilon_v^2}},\\
\Delta\mathbf{P}&=m\hat{\mathbf{v}} .
\end{aligned}
\label{eq:flow_param}
\end{equation}
Here $\tilde{\mathbf{v}}\in\mathbb{R}^3$ and $\tilde{m}\in\mathbb{R}$ are predicted from $\mathbf{q}_c$ and $\epsilon_v=10^{-4}$. Because of the stabilizing $\epsilon_v$, $\hat{\mathbf v}$ is not constrained to have exactly unit norm. This four-parameter representation separates motion orientation from motion magnitude, guarantees a non-negative magnitude, and lets the two factors receive complementary supervision.

For scene-flow supervision, let $\Delta\mathbf{P}^{*}$ denote the ground-truth displacement and define the stable radial log transform
\begin{equation}
\phi(\mathbf{x})=
\mathbf{x}\frac{\log(1+\|\mathbf{x}\|_{\epsilon})}{\|\mathbf{x}\|_{\epsilon}},
\qquad
\|\mathbf{x}\|_{\epsilon}=\sqrt{\|\mathbf{x}\|_2^2+\epsilon_{\Delta}^{\,2}},
\label{eq:flow_log}
\end{equation}
where $\epsilon_{\Delta}=10^{-6}$. This transform preserves the displacement direction while compressing its dynamic range, preventing large motions from overwhelming small ones. We classify a valid query as moving when the metric ground-truth magnitude exceeds $\gamma=0.02$\,m, yielding dynamic and static sets $\mathcal{D}$ and $\mathcal{S}$. The displacement term is
\begin{equation}
\begin{aligned}
\ell_i^\Delta
&=\|\phi(\Delta\mathbf{P}_i)-\phi(\Delta\mathbf{P}^{*}_i)\|_1,\\
\mathcal{L}_{\Delta}
&=\alpha\,\underset{i\in\mathcal{D}}{\mathrm{mean}}\,\ell_i^\Delta
+(1-\alpha)\,\underset{i\in\mathcal{S}}{\mathrm{mean}}\,\ell_i^\Delta .
\end{aligned}
\label{eq:flow_disp_loss}
\end{equation}
Here $\alpha$ controls the relative emphasis on moving points; we use $\alpha=0.85$ during training. If one set is empty, only the available term is used. We further supervise orientation on moving points and suppress spurious motion on static points:
\begin{equation}
\begin{aligned}
\hat{\mathbf{v}}_i^{*}
&=\Delta\mathbf{P}^{*}_i/\|\Delta\mathbf{P}^{*}_i\|_2,\\
\mathcal{L}_{\mathrm{dir}}
&=\underset{i\in\mathcal{D}}{\mathrm{mean}}
\left(1-\hat{\mathbf{v}}_i^\top\hat{\mathbf{v}}_i^{*}\right)
+\lambda_{\mathrm{stat}}\underset{i\in\mathcal{S}}{\mathrm{mean}}\,m_i,
\quad \lambda_{\mathrm{stat}}=1 .
\end{aligned}
\label{eq:flow_dir_loss}
\end{equation}
Thus, the composed displacement is constrained globally in log space, while its direction and near-zero static magnitude are each explicitly regularized.

\subsection{Training Objectives}
\label{sec:training}

UniQuery4R is trained with a multi-task objective that supervises the full set of query outputs and camera parameters:
\begin{equation}
\begin{aligned}
\mathcal{L}
&=
\lambda_{P}\,\mathcal{L}_{P}
+
\lambda_{\Delta}\,\mathcal{L}_{\Delta}
+
\lambda_{f}\,\mathcal{L}_{f}
+
\lambda_{\mathrm{dir}}\,\mathcal{L}_{\mathrm{dir}}
\\
&\quad
+
\lambda_{d}\,\mathcal{L}_{d}
+
\lambda_{\pi}\,\mathcal{L}_{\pi},
\end{aligned}
\label{eq:loss}
\end{equation}
where $\mathcal{L}_{P}$ and $\mathcal{L}_{d}$ are confidence-weighted $L_1$ losses on the 3D point and source depth; $\mathcal{L}_{\Delta}$ balances static and dynamic regions; $\mathcal{L}_{\mathrm{dir}}$ supervises moving-point direction and suppresses the magnitude on static points; $\mathcal{L}_{f}$ is a confidence-aware 2D matching loss; and $\mathcal{L}_{\pi}$ supervises reference-normalized camera parameters. Dataset-dependent reweighting disables unavailable supervision and reduces the point-loss weight on sparse-reconstruction data. Additional implementation details are provided in the supplementary material.

\section{Experiments}

% Main WorldTrack tables + Figure 3 on one results page. Figure height is
% tuned just below the split threshold so the three floats fill the page.
\begin{figure*}[p]
  \centering
  \small
  \begin{tabular}{@{}l@{\hskip4pt}*{9}{c@{\hskip6pt}}c@{}}
    \toprule
    Method & \multicolumn{2}{c}{PStudio} & \multicolumn{2}{c}{PO} & \multicolumn{2}{c}{DR} & \multicolumn{2}{c}{ADT} & \multicolumn{2}{c}{Macro Avg.} \\
    \cmidrule(lr){2-3} \cmidrule(lr){4-5} \cmidrule(lr){6-7} \cmidrule(lr){8-9} \cmidrule(lr){10-11}
    & $\tau\uparrow$ & EPE$\downarrow$ & $\tau\uparrow$ & EPE$\downarrow$ & $\tau\uparrow$ & EPE$\downarrow$ & $\tau\uparrow$ & EPE$\downarrow$ & $\tau\uparrow$ & EPE$\downarrow$ \\
    \midrule
    SpatialTrackerV2~\cite{xiao2025spatialtrackerv2} & 8.78 & 0.4095 & 15.08 & 0.3089 & 13.09 & 0.2707 & 34.40 & 0.2054 & 17.84 & 0.2986 \\
    St4RTrack~\cite{feng2025st4rtrack}             & 27.45 & 0.2259 & 37.24 & 0.219 & 75.69 & 0.0858 & 55.76  & 0.1481 & 49.04 & 0.1697 \\
    TraceAnything~\cite{liu2025traceanything}      & 23.94 & 0.2456 & 64.33 & 0.1173 & 83.83 & 0.0495 & 79.98  & 0.0839 & 63.02 & 0.1241 \\
    Any4D~\cite{karhade2025any4d}                  & 25.08 & 0.2406 & 70.43 & 0.094 & 85.94 & 0.0451 & 88.25  & 0.0562 & 67.43 & 0.109 \\
    V-DPM~\cite{sucar2026vdpm}                     & \textbf{58.99} & \underline{0.1152} & \underline{82.91} & 0.0563 & 88.98 & 0.0381 & \underline{88.75} & 0.0559 & 79.91 & 0.0664 \\
    4RC~\cite{luo20264rc}                          & 53.88 & 0.1255 & \textbf{86.66} & \textbf{0.0466} & \textbf{94.13} & \textbf{0.0213} & 88.14  & 0.0633 & \underline{80.70} & \underline{0.0642} \\
    OpenD4RT~\cite{opend4rt}                       & 38.25 & 0.1772 & 76.62 & 0.0784 & 90.45 & 0.0289 & 88.23  & 0.0627 & 73.39 & 0.0868 \\
    \midrule
    \textbf{UniQuery4R (ours)}                     & \underline{56.75} & \textbf{0.1137} & 82.52 & \underline{0.0541} & \underline{91.54} & \underline{0.0275} & \textbf{95.10} & \textbf{0.0167} & \textbf{81.48} & \textbf{0.0530} \\
    \bottomrule
  \end{tabular}
  \captionof{table}{WorldTrack scene flow~\cite{feng2025st4rtrack}: source\,=\,first
  frame; median global scale alignment; $\tau@0.1\mathrm{m}$ (\%) and EPE (m);
  equal-weight macro-average. OpenD4RT: unofficial D4RT~\cite{opend4rt}.
  Best bolded; second-best \underline{underlined}.}
  \label{tab:scene_flow}

  \centering
  \small
  \begin{tabular}{@{}l@{\hskip4pt}*{9}{c@{\hskip6pt}}c@{}}
    \toprule
    Method & \multicolumn{2}{c}{PStudio} & \multicolumn{2}{c}{PO} & \multicolumn{2}{c}{DR} & \multicolumn{2}{c}{ADT} & \multicolumn{2}{c}{Macro Avg.} \\
    \cmidrule(lr){2-3} \cmidrule(lr){4-5} \cmidrule(lr){6-7} \cmidrule(lr){8-9} \cmidrule(lr){10-11}
    & APD$\uparrow$ & EPE$\downarrow$ & APD$\uparrow$ & EPE$\downarrow$ & APD$\uparrow$ & EPE$\downarrow$ & APD$\uparrow$ & EPE$\downarrow$ & APD$\uparrow$ & EPE$\downarrow$ \\
    \midrule
    SpatialTrackerV2~\cite{xiao2025spatialtrackerv2} & 49.93 & 0.4639 & 53.45 & 0.5173 & 54.34 & 0.4368 & 69.39 & 0.2764 & 56.78 & 0.4236 \\
    St4RTrack~\cite{feng2025st4rtrack}             & 70.80 & 0.2479 & 67.38 & 0.314 & 74.24 & 0.2594 & 72.90 & 0.2958 & 71.33 & 0.2793 \\
    TraceAnything~\cite{liu2025traceanything}      & 70.86 & 0.278 & 39.37 & 1.0462 & 60.89 & 0.5642 & 75.69 & 0.2515 & 61.70 & 0.535 \\
    Any4D~\cite{karhade2025any4d}                  & 74.72 & 0.2361 & 65.99 & 0.3783 & 76.72 & 0.3463 & 78.09 & 0.2378 & 73.88 & 0.2996 \\
    V-DPM~\cite{sucar2026vdpm}                     & \underline{78.95} & \underline{0.1757} & \underline{80.82} & \underline{0.1857} & 75.29 & 0.2438 & \underline{85.47} & \underline{0.1725} & \underline{80.13} & \underline{0.1944} \\
    4RC~\cite{luo20264rc}                          & 69.08 & 0.259 & 79.81 & 0.2353 & \textbf{84.65} & \textbf{0.1667} & 84.16 & 0.177 & 79.43 & 0.2095 \\
    OpenD4RT~\cite{opend4rt}                       & 78.63 & 0.1811 & 66.03 & 0.3398 & 76.78 & 0.2407 & 69.91 & 0.2966 & 72.84 & 0.2646 \\
    \midrule
    \textbf{UniQuery4R (ours)}                     & \textbf{84.64} & \textbf{0.1333} & \textbf{83.74} & \textbf{0.1604} & \underline{80.00} & \underline{0.2078} & \textbf{87.17} & \textbf{0.1387} & \textbf{83.89} & \textbf{0.1601} \\
    \bottomrule
  \end{tabular}
  \captionof{table}{WorldTrack dynamic-point tracking~\cite{feng2025st4rtrack}: queried
  source-point 3D positions (64 frames $\times$ 50 sequences); median global
  scale alignment; APD (\% over $0.1/0.3/0.5/1.0$\,m) and EPE (m);
  equal-weight macro-average. OpenD4RT: unofficial D4RT~\cite{opend4rt}.
  Best bolded; second-best \underline{underlined}.}
  \label{tab:dynamic_points}

  \centering
  \includegraphics[width=0.96\textwidth,keepaspectratio]{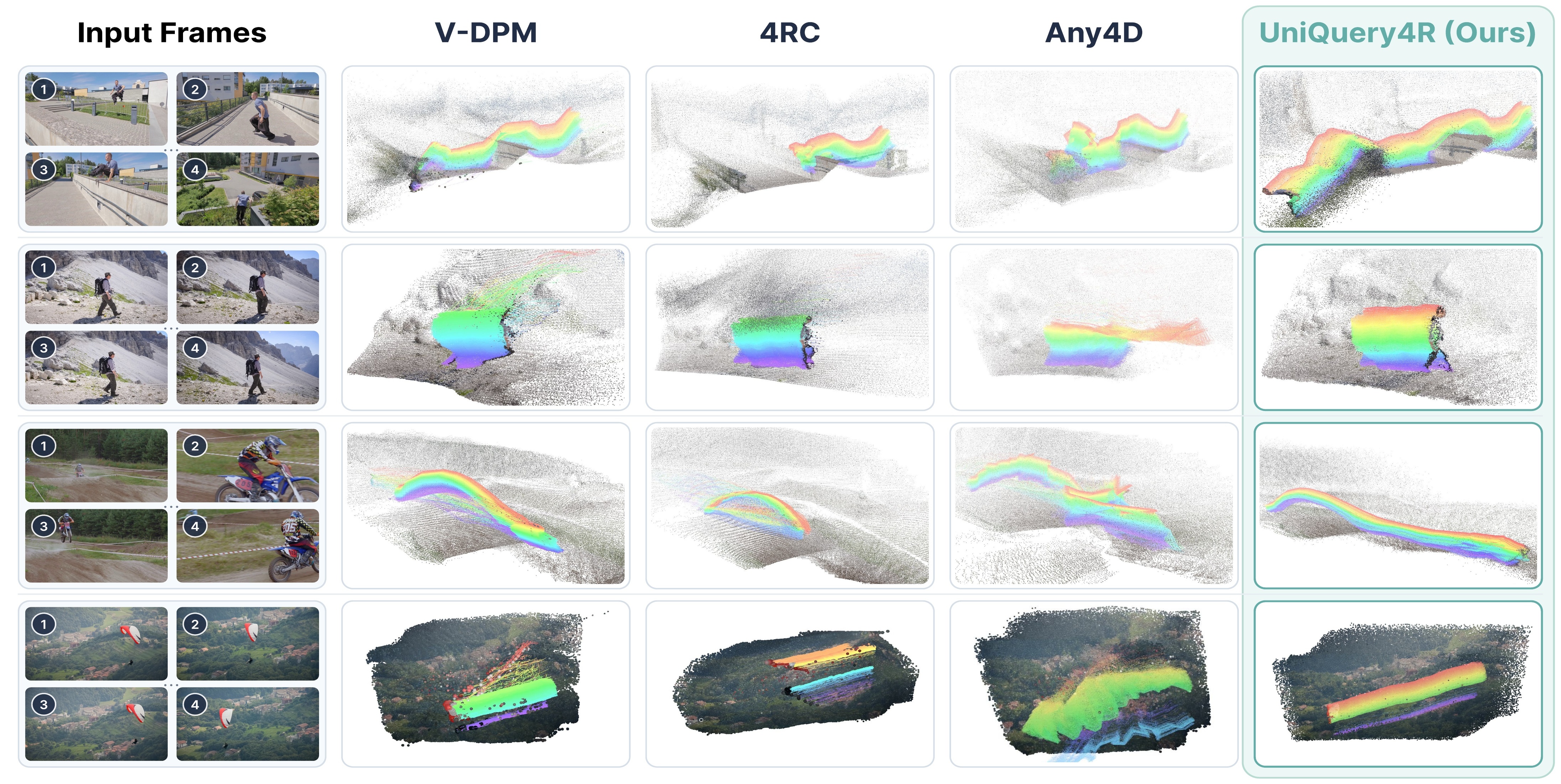}
  \captionof{figure}{Long-sequence DAVIS tracking versus feed-forward baselines.
  Each color denotes one fixed source query over time; UniQuery4R
  qualitatively exhibits less drift and fragmentation under occlusion and
  deformation.}
  \label{fig:davis_comparison}
\end{figure*}

\subsection{Training Data and Setup}
\label{sec:training_data}
UniQuery4R is trained on a mixture of synthetic and real-world dynamic datasets, including Kubric-4D~\cite{vanhoorick2024gcd,greff2022kubric}, PointOdyssey~\cite{zheng2023pointodyssey}, Hypersim~\cite{roberts2021hypersim}, DL3DV~\cite{ling2024dl3dv}, CoTracker3Kubric~\cite{karaev2025cotracker3}, Stereo4D~\cite{jin2025stereo4d}, Virtual KITTI~2~\cite{cabon2020vkitti2}, Waymo~\cite{sun2020waymo}, and an internal dynamic-scene collection. Together they supervise depth, camera pose, optical flow, scene flow, and 2D/3D trajectories. All data follow a unified coordinate convention. Training clips contain $4$--$12$ views with varied baselines and motion magnitudes; for each clip, we sample ordered source--target pairs $(s,t)$, including self-pairs, for query supervision. The main model is trained for $150$k iterations on $16$ NVIDIA H20 GPUs; optimizer settings and full hyperparameters are in the supplementary material.

\subsection{Evaluation Protocol}
\label{sec:metrics_geom_accuracy}
We follow the WorldTrack protocol~\cite{feng2025st4rtrack} on
PointOdyssey (PO)~\cite{zheng2023pointodyssey}, Panoptic Studio
(PStudio)~\cite{joo2015panoptic}, Dynamic Replica
(DR)~\cite{karaev2023dynamicstereo}, and Aria Digital Twin
(ADT)~\cite{pan2023aria}, with qualitative long-sequence tracking on
DAVIS~\cite{perazzi2016davis}. Quantitative runs use the first $64$
frames with global median scale alignment before metric computation;
macro-averages are unweighted over the four datasets. Table captions
detail source--target sampling, validity masks, and thresholds.
Although trained on $4$--$12$-frame clips, UniQuery4R jointly encodes
all $64$ evaluation frames in one encoder pass, and source--target
queries reuse those features without sliding windows or cross-window
fusion. All entries labeled D4RT use OpenD4RT~\cite{opend4rt}, an
unofficial reimplementation, and should not be read as results of the
original model~\cite{zhang2025d4rt}.

\subsection{Scene Flow Evaluation}
\label{sec:scene_flow_results}
Table~\ref{tab:scene_flow} reports $\tau@0.1\mathrm{m}$ and EPE with
unweighted four-dataset macro-averages. UniQuery4R attains the best
macro-average on both metrics and the best ADT scores, ranks second
on DR, and trails V-DPM/4RC on individual PStudio/PO
dataset--metric combinations. The direction--magnitude design in
Sec.~\ref{sec:scene_flow_modeling} balances static/dynamic supervision
so large displacements do not dominate training
(Table~\ref{tab:ablation}).

\subsection{Dynamic Point Evaluation}
\label{sec:dynamic_points_results}
Table~\ref{tab:dynamic_points} evaluates world-coordinate 3D tracking
of queried dynamic source points via Average Percentage of Points
within Distance (APD) and EPE after global median scale alignment.
UniQuery4R leads the four-dataset macro-average and ranks first on
PStudio, PO, and ADT, while 4RC is strongest on DR. The dynamic-point
and scene-flow heads decode from the shared query representation
$\mathbf{q}_c$, providing a common representation for correspondence,
geometry, and motion. Figure~\ref{fig:davis_comparison} qualitatively
shows less drift and fragmentation under occlusion and non-rigid
deformation.
\begin{table*}[t]
  \centering
  \small
  \setlength{\tabcolsep}{1.6pt}
  \begin{tabular}{@{}l*{15}{c}@{}}
  \toprule
  Method
  & \multicolumn{5}{c}{ADT}
  & \multicolumn{5}{c}{NRGBD}
  & \multicolumn{5}{c}{Sintel} \\
  \cmidrule(lr){2-6}\cmidrule(lr){7-11}\cmidrule(lr){12-16}
  & A$_5$$\uparrow$ & A$_{30}$$\uparrow$ & ATE$\downarrow$ & R$_t$$\downarrow$ & R$_r$$\downarrow$
  & A$_5$$\uparrow$ & A$_{30}$$\uparrow$ & ATE$\downarrow$ & R$_t$$\downarrow$ & R$_r$$\downarrow$
  & A$_5$$\uparrow$ & A$_{30}$$\uparrow$ & ATE$\downarrow$ & R$_t$$\downarrow$ & R$_r$$\downarrow$ \\
  \midrule
  VGGT~\cite{wang2025vggt}
  & 0.134 & 0.724 & 0.219 & 0.269 & 3.457
  & 0.676 & 0.926 & 0.047 & 0.071 & 0.994
  & 0.212 & 0.475 & 0.100 & 0.050 & 0.588 \\
  DA3-Giant~\cite{lin2025depthanything3}
  & 0.145 & 0.741 & 0.160 & 0.223 & 2.221
  & \textbf{0.772} & \underline{0.944} & \underline{0.034} & \textbf{0.045} & \textbf{0.848}
  & 0.214 & 0.495 & 0.065 & \underline{0.044} & \underline{0.373} \\
  VGGT-$\Omega$~\cite{wang2026vggtomega}
  & 0.446 & 0.851 & 0.087 & 0.091 & 2.107
  & 0.704 & 0.933 & 0.040 & 0.059 & 1.008
  & \underline{0.286} & \underline{0.583} & \textbf{0.028} & \textbf{0.022} & \textbf{0.230} \\
  \midrule
  Any4D~\cite{karhade2025any4d}
  & 0.070 & 0.597 & 0.346 & 0.432 & 4.566
  & 0.348 & 0.833 & 0.105 & 0.163 & 2.247
  & 0.034 & 0.290 & 0.144 & 0.104 & 2.226 \\
  V-DPM~\cite{sucar2026vdpm}
  & 0.400 & 0.861 & 0.076 & 0.103 & 2.374
  & 0.664 & 0.925 & 0.045 & 0.066 & 1.113
  & 0.175 & 0.505 & 0.082 & 0.054 & 0.520 \\
  4RC~\cite{luo20264rc}
  & \underline{0.571} & \textbf{0.906} & \underline{0.046} & \textbf{0.064} & \underline{1.647}
  & \underline{0.753} & 0.942 & 0.035 & \underline{0.050} & \underline{0.932}
  & 0.240 & 0.488 & 0.089 & 0.046 & 0.493 \\
  \midrule
  \textbf{UniQuery4R (ours)}
  & \textbf{0.608} & \underline{0.903} & \textbf{0.036} & \underline{0.079} & \textbf{1.446}
  & 0.708 & \textbf{0.951} & \textbf{0.031} & 0.080 & 1.460
  & \textbf{0.359} & \textbf{0.603} & \underline{0.031} & 0.059 & 1.199 \\
    \bottomrule
  \end{tabular}
  \caption{Camera pose (A$_5$/A$_{30}$: AUC@5/@30) and trajectory (ATE;
  R$_t$/R$_r$: RPE$_t$/RPE$_r$) on ADT, NRGBD, and Sintel. Best bolded;
  second-best underlined.}
  \label{tab:camera_evaluation}
\end{table*}

\subsection{Depth and Camera Evaluation}
\label{sec:depth_camera_results}

\noindent
\begin{minipage}{\columnwidth}
  \centering
  \small
  \setlength{\tabcolsep}{2.4pt}
  \begin{tabular}{@{}l*{6}{c}@{}}
  \toprule
  Method
  & \multicolumn{2}{c}{ADT}
  & \multicolumn{2}{c}{Sintel}
  & \multicolumn{2}{c}{ScanNet++} \\
  \cmidrule(lr){2-3}\cmidrule(lr){4-5}\cmidrule(lr){6-7}
  & Rel$\downarrow$ & $\delta$$\uparrow$
  & Rel$\downarrow$ & $\delta$$\uparrow$
  & Rel$\downarrow$ & $\delta$$\uparrow$ \\
  \midrule
  VGGT
  & 0.0814 & 0.9286
  & 0.2420 & 0.6747
  & 0.0379 & \underline{0.9765} \\
  DA3-Giant
  & 0.0649 & 0.9735
  & 0.2906 & 0.6357
  & \underline{0.0343} & 0.9757 \\
  VGGT-$\Omega$
  & 0.0638 & \underline{0.9763}
  & \textbf{0.0982} & \textbf{0.9163}
  & 0.0379 & 0.9727 \\
  \midrule
  Any4D
  & 0.0999 & 0.9553
  & 0.2514 & 0.6468
  & 0.0614 & 0.9666 \\
  V-DPM
  & \underline{0.0564} & 0.9727
  & 0.2207 & 0.6819
  & 0.0461 & 0.9678 \\
  4RC
  & 0.0681 & 0.9514
  & 0.2388 & 0.6608
  & \textbf{0.0255} & \textbf{0.9789} \\
  \midrule
  \textbf{UniQuery4R}
  & \textbf{0.0433} & \textbf{0.9769}
  & \underline{0.1033} & \underline{0.8689}
  & 0.0549 & 0.9622 \\
    \bottomrule
  \end{tabular}
  \captionof{table}{Video depth evaluation on ADT, Sintel, and ScanNet++ (AbsRel
  (Rel)$\downarrow$, $\delta_{1.25}$ ($\delta$)$\uparrow$). Best bolded;
  second-best underlined.}
  \label{tab:depth_evaluation}
\end{minipage}

Tables~\ref{tab:camera_evaluation} and~\ref{tab:depth_evaluation} summarize
depth and camera results. UniQuery4R leads ADT depth and ADT AUC@5/ATE/RPE$_r$;
4RC is stronger on AUC@30/RPE$_t$. On Sintel, AbsRel is close to
VGGT-$\Omega$ with stronger pose AUC but weaker trajectory RPE. On NRGBD,
it leads AUC@30 and ATE but trails DA3-Giant/4RC in relative RPE; on
ScanNet++, depth remains below 4RC and reconstruction-oriented backbones.

\subsection{Runtime and Memory Scaling}
\label{sec:runtime_scaling}
\noindent
\begin{minipage}{\columnwidth}
  \centering
  \small
  \setlength{\tabcolsep}{3pt}
  \begin{tabular}{@{}llc@{}}
\toprule
Component & Scaling variable & Marginal latency \\
\midrule
Encoder & One additional view & ${+}45.70$\,ms \\
Encoder & ${+}1$\,MP per clip & ${+}1.26$\,s \\
Decoder & ${+}1$k queries & ${+}4.18$\,ms \\
    \bottomrule
  \end{tabular}
  \captionof{table}{FP16 inference scaling on one NVIDIA A800
  (warmup~$5$, repeats~$20$). Defaults: $504{\times}504$, $Q{=}4096$;
  view sweep fixes $504^2$/$Q{=}4096$, resolution sweep fixes $T{=}8$/$Q{=}4096$,
  query sweep fixes $T{=}8$/$504^2$. Linear fits give $R^2{\ge}0.995$.
  Encoding changes by only $0.22\%$ for $Q{:}4096{\to}16384$. At $Q{=}60$k,
  chunking ($30$k) cuts peak memory $15.68{\to}12.52$\,GB. Full protocol in
  the supplementary material.}
  \label{tab:runtime_scaling}
\end{minipage}

Table~\ref{tab:runtime_scaling} summarizes encode/decode marginal cost:
encoding grows with views and resolution and is nearly $Q$-independent,
while decoding scales with query count; chunking reduces peak memory at
large $Q$. Detailed sweeps are in the supplementary material.

\subsection{Ablation Study}
\label{sec:ablation}
\noindent
\begin{minipage}{\columnwidth}
  \centering
  \small
  \setlength{\tabcolsep}{1.2mm}
  \begin{tabular}{@{}l@{\hskip1.2mm}c@{\hskip1.2mm}c@{\hskip1.2mm}c@{\hskip1.2mm}c@{}}
  \toprule
  \multicolumn{2}{@{}l}{Supervised tasks} & APD$\uparrow$ & Point $\tau\uparrow$ & Point EPE$\downarrow$ \\
  \midrule
  \multicolumn{2}{@{}l}{Dynamic points + scene flow}       & 77.66 & 39.56 & 0.2161 \\
  \multicolumn{2}{@{}l}{\quad + 2D correspondence}         & 78.22 & 42.04 & 0.2109 \\
  \multicolumn{2}{@{}l}{\quad + local depth (ref.)}  & \textbf{78.78} & \textbf{42.93} & \textbf{0.2060} \\
  \midrule
  \multicolumn{2}{@{}l}{Flow parameterization} & Flow $\tau\uparrow$ & Flow EPE$\downarrow$ & \\
  \multicolumn{2}{@{}l}{Cartesian vector $(d_x,d_y,d_z)$} & 73.71 & 0.0794 & \\
  \multicolumn{2}{@{}l}{Direction + magnitude}    & \textbf{76.48} & \textbf{0.0663} & \\
  \midrule
  Source pyramid & CA blocks & APD$\uparrow$ & Point $\tau\uparrow$ & Point EPE$\downarrow$ \\
  w/o multi-scale & 4 & 74.71 & 35.64 & 0.2531 \\
  multi-scale & 1 & 76.78 & 39.39 & 0.2255 \\
  multi-scale & 2 & 76.89 & 39.77 & 0.2261 \\
  multi-scale (ref.) & 4 & \textbf{78.78} & \textbf{42.93} & \textbf{0.2060} \\
    \bottomrule
  \end{tabular}
  \captionof{table}{WorldTrack ablations (macro-average; $75$k / one H20).
  Best per group bolded; ref.\ is multi-scale + 4 CA.}
  \label{tab:ablation}
\end{minipage}

Table~\ref{tab:ablation} shows that correspondence and local depth improve
dynamic-point metrics; direction--magnitude raises flow
$\tau@0.1\mathrm{m}$ from $73.71$ to $76.48$ and lowers flow EPE from
$0.0794$ to $0.0663$, with four multi-scale CA blocks performing best.

\subsection{Discussion and Limitations}
Figure~\ref{fig:davis_comparison} suggests improved long-horizon stability in
the shown examples relative to feed-forward baselines, and
direction--magnitude supervision is central (Table~\ref{tab:ablation}).
Performance can still degrade under severe occlusion and large rotations;
with fast motion or flipping articulations, left--right hand/foot
associations are particularly prone to identity swaps. Encode cost also
grows with clip length (Table~\ref{tab:runtime_scaling}), so longer
videos need temporal windowing; cross-window fusion is left open.

\section{Conclusion}
UniQuery4R queries continuous source pixels over a jointly encoded clip with
direction--magnitude scene flow (Sec.~\ref{sec:scene_flow_modeling}),
achieving the best WorldTrack macro-average scene-flow and dynamic-point
results among evaluated methods, with competitive depth and camera metrics. Ablations confirm gains from correspondence, local depth,
direction--magnitude flow, and a multi-scale source pyramid.

\bibliography{references}

\setcounter{section}{0}
\setcounter{table}{0}
\setcounter{figure}{0}
\setcounter{equation}{0}
\makeatletter
\@addtoreset{table}{section}
\@addtoreset{figure}{section}
\@addtoreset{equation}{section}
\makeatother
\renewcommand{\thesection}{\Alph{section}}
\renewcommand{\thesubsection}{\thesection.\arabic{subsection}}
\renewcommand{\thetable}{\thesection.\arabic{table}}
\renewcommand{\thefigure}{\thesection.\arabic{figure}}
\renewcommand{\theequation}{\thesection.\arabic{equation}}

\twocolumn[
\begin{center}
{\LARGE\bfseries UniQuery4R: Unified 4D Scene Reconstruction from a Single Query}

\vspace{12pt}
{\Large\bfseries Supplementary Material}

\end{center}
]

\section{Implementation Details}
\label{sec:supp-impl}

\subsection{Coordinate Frame and Scale}
\label{sec:supp-scale}

All 3D outputs and camera translations use the view-$0$ reference
frame. During training, geometry is normalized by the clip-level mean
point-to-origin distance
\begin{equation}
a=\frac{1}{|\Omega|}\sum_{(n,p)\in\Omega}\bigl\|\mathbf X_{n,p}^{(0)}\bigr\|_2,
\end{equation}
computed over valid depth pixels $\Omega$; metric labels are restored by
multiplying by~$a$. At evaluation, WorldTrack applies median-based
global scale alignment separately to each output type:
\begin{equation}
s=\frac{\mathrm{median}(\|\mathbf y^*\|)}{\mathrm{median}(\|\hat{\mathbf y}\|)}.
\end{equation}
Camera losses average relative-pose terms over all reference views.

\subsection{Multi-View Backbone}
\label{sec:supp-backbone}

The encoder is an internal multi-view ViT-G ($1536$ width, $40$ blocks,
$14{\times}14$ patches) with joint multi-view attention, initialized from
DA3-Giant~\cite{lin2025depthanything3} and further pretrained on an
internal mixture (corpus and schedule not released).
Section~\ref{sec:supp-backbone-init} compares alternative initializations.
During UniQuery4R training the backbone uses $0.1\times$ the learning rate
of newly added modules. We extract $3072$-D tokens from blocks
$\{19,27,33,39\}$, project each stage, and PixelShuffle with factors
$(4,2,1,1)$ to build the four-level pyramid fed to the query decoder.
Rotary positional encoding and query/key normalization start at block~$13$.

\subsection{Query Decoder}

The query decoder operates at width $256$. Source coordinates are mapped
to $[-1,1]^2$ for continuous bilinear sampling; up to $100$k queries are
supported per forward pass. We use four Pre-LN source--target cross-attention
blocks ($8$ heads, head width $32$, FFN${\times}4$, GELU, no dropout).
At training time, queries follow dataset annotations when available and
otherwise use a dense integer grid; at inference, queries may be placed
continuously on the source image.

\subsection{Prediction Heads}
\label{sec:supp-heads}

Two-layer ReLU MLPs (hidden $64$) decode the outputs listed in
Table~\ref{tab:supp_heads}.
\begin{table}[!t]
\centering
\setlength{\tabcolsep}{4pt}
\begin{tabular}{@{}llccc@{}}
\toprule
Output & From & Ch. & Act. & Notes \\
\midrule
$\mathbf P$ & $\mathbf q_c$ & 3 & inv-log & target-time 3D \\
$c_P$ & $\mathbf q_c$ & 1 & --- & logit \\
$\tilde{\mathbf v},\tilde m$ & $\mathbf q_c$ & $3{+}1$ & dir.--mag. & $\to\Delta\mathbf P$ \\
$\mathbf f$ & $\mathbf q_c$ & 2 & identity & abs.\ UV \\
$\mathbf c_f$ & $\mathbf q_c$ & 4 & --- & logits \\
$d,c_d$ & $\mathbf q_s$ & $1{+}1$ & $\exp$/--- & source depth \\
\bottomrule
\end{tabular}
\caption{Query prediction heads.}
\label{tab:supp_heads}
\end{table}
\noindent
Target-time 3D points and source depth use
\begin{equation}
\mathbf P=\mathrm{sign}(\tilde{\mathbf P})\odot\bigl(e^{|\tilde{\mathbf P}|}-1\bigr),
\qquad
d=\exp(\tilde d).
\end{equation}

\subsection{Camera Head}

Following VGGT~\cite{wang2025vggt}, per-view camera tokens pass through a
four-block, $16$-head trunk (width $3072$) with four additive pose updates.
Supervision follows $\pi^3$~\cite{wang2026pi3} relative-pose $L_1$ on
translation, quaternion ($w{\ge}0$), and FoV, averaged over reference
views; the principal point is fixed at the image center.

\subsection{Losses}
\label{sec:supp-objectives}

The full objective combines $\mathcal L_P$, $\mathcal L_d$, $\mathcal L_f$,
$\mathcal L_\pi$, scene-flow terms, and dataset-dependent reweighting.
Table~\ref{tab:supp_loss_weights} lists the verified weights used for the
final model (outer multipliers: depth/motion $=1$, camera $=10$).
\begin{table*}[!t]
\centering
\setlength{\tabcolsep}{4pt}
\begin{tabular*}{\textwidth}{@{\extracolsep{\fill}}lcl@{}}
\toprule
Term & Default & Dataset overrides\\
\midrule
Source depth $\mathcal L_d$ & $1$ &
DL3DV/Waymo: $0$; VKITTI2/Stereo4D: $0$\\
Target point $\mathcal L_P$ & $10$ &
DL3DV: $.1$; Waymo/VKITTI2: $0$; Stereo4D: $.1$\\
Displacement $\mathcal L_\Delta$ & $50$ &
Waymo/VKITTI2/Stereo4D: $30$\\
Direction/static magnitude & $1$ & none\\
2D correspondence $\mathcal L_f$ & $5$ & PointOdyssey: $0$\\
Camera $\mathcal L_\pi$ & $1$ & outer multiplier $10$\\
\bottomrule
\end{tabular*}
\caption{Loss weights used to train the final model. Camera translation,
quaternion, and FoV components have internal weights $1$, $10$, and
$0.5$, respectively. The four refinement predictions are weighted by
$\{0.6^3,0.6^2,0.6,1\}$ and then averaged.}
\label{tab:supp_loss_weights}
\end{table*}

\paragraph{Depth and 3D point.}
Confidence-weighted $L_1$ residuals are
\begin{equation}
r_d=\frac{|d-d^*|}{|d^*|+10^{-6}},
\qquad
r_P=\frac{\|\mathbf P-\mathbf P^*\|_1}{|P_z^*|+10^{-6}},
\end{equation}
with outliers above $3$ discarded (and the top $1\%$ when more than $1000$
valid points remain). With logit~$z$, confidence $c=1+\exp(z)$ enters as
$r_{\mathrm{raw}}c-0.1\log c$.

\paragraph{2D correspondence.}
The matching error is $e=\|\mathbf f-\mathbf f^*\|_2$ in normalized
$[-1,1]^2$, combined with a robust kernel
\begin{equation}
\rho(e)=c_s^\alpha\Bigl(\bigl(e/c_s\bigr)^2+1\Bigr)^{\alpha/2},
\qquad \alpha=0.5,\; c_s=10^{-4},
\end{equation}
plus visibility and one-pixel BCE terms (weight $0.01$); source-valid but
target-occluded tracks keep weight $0.2$.

\paragraph{Scene flow.}
Supervision uses the direction--magnitude parameterization with separate
dynamic and static terms; transformed residuals above $10$ are rejected.

\subsection{Data and Optimization}
\label{sec:supp-data}

Table~\ref{tab:supp_data_mix} lists the training mixture (relative
weights need not sum to one).
\begin{table}[!t]
\centering
\setlength{\tabcolsep}{4pt}
\begin{tabular}{@{}lrr@{}}
\toprule
Source & \#samples & Weight\\
\midrule
PointOdyssey & 111 & 0.4\\
Kubric-4D & 12,928 & 1.0\\
CoTracker3Kubric & 5,827 & 1.0\\
Internal dataset & 2,596 & 1.0\\
Hypersim & 758 & 0.1\\
DL3DV & 54,903 & 0.1\\
Waymo & 5,497 & $1/15$\\
Virtual KITTI 2 & 1,144 & $1/15$\\
Stereo4D & 163,924 & $1/15$\\
\bottomrule
\end{tabular}
\caption{Training mixture (indexed samples and relative weights).}
\label{tab:supp_data_mix}
\end{table}
\noindent
Clips use $4$--$12$ temporally sorted frames; for an $N$-view clip we
sample $N$ ordered $(s,t)$ pairs including self-pairs. SynthVerse is
excluded from the main zero-shot checkpoint. Images keep aspect ratio
(max side $504$), normalize by $127.5$, and use photometric / blur /
compression augmentation without horizontal flip.

Table~\ref{tab:supp_main_training} lists the optimizer, learning-rate
schedule, and related hyperparameters of the final 16-GPU run.
\begin{table}[!t]
\centering
\setlength{\tabcolsep}{4pt}
\begin{tabular}{@{}lc@{}}
\toprule
Setting & Value\\
\midrule
Hardware & $16\times$ NVIDIA H20\\
Iterations & $150$k\\
Numerical precision & bfloat16\\
Batch size (per GPU / global) & $1$ / $16$\\
Backbone learning rate & $10^{-5}$\\
Other-module learning rate & $10^{-4}$\\
AdamW $(\beta_1,\beta_2)$ & $(0.9,0.999)$\\
AdamW $\epsilon$ / weight decay & $10^{-10}$ / $10^{-3}$\\
Warmup & 1k iterations, linear\\
Schedule & cosine decay to $10^{-8}$ (base)\\
Max image side & $504$\\
Gradient clipping & global norm $1$\\
Random seed & 2024\\
\bottomrule
\end{tabular}
\caption{Optimization hyperparameters for the final model.}
\label{tab:supp_main_training}
\end{table}
\noindent
Ablations use 75k iterations on one H20 (global batch $1$) under a
matched half-schedule; absolute numbers are not comparable to the final
150k model.

\noindent\textbf{Reproducibility scope.}
The backbone and part of the training data are internal.

\FloatBarrier
\section{Additional Experiments}
\label{sec:supp-extra-results}

\noindent\textbf{Evaluation protocol.}
WorldTrack reports an equal-weight four-dataset macro-average with the
first frame as source. OpenD4RT is an unofficial reimplementation.
Flow4R is omitted because no public code or weights are available.

\subsection{Runtime and Memory Scaling}
\label{sec:supp-runtime}

Table~\ref{tab:runtime_raw} lists the measured runtime points
underlying the scaling analysis.
\begin{table*}[t]
\centering
\small
\setlength{\tabcolsep}{3pt}
\begin{tabular}{rrcrccrrrr}
\toprule
\multicolumn{2}{c}{Encode vs.\ clip length}
&
\multicolumn{2}{c}{Encode vs.\ resolution}
&
\multicolumn{6}{c}{Encode/decode vs.\ query count}
\\
\cmidrule(lr){1-2}
\cmidrule(lr){4-5}
\cmidrule(lr){7-10}
$T$
& Enc.\ (ms)
&
& $H{\times}W$
& Enc.\ (ms)
&
& Mode
& $Q$ (chunks)
& Enc.\ (ms)
& Query dec.\ (ms)
\\
\midrule
2  & $118.90{\pm}14.48$ &&
$252^2$ & $107.62{\pm}13.36$ &&
S & 256 (1)       & 357.79 & $5.36{\pm}0.10$ \\

4  & $200.06{\pm}10.73$ &&
$364^2$ & $197.37{\pm}5.29$ &&
S & 1,024 (1)     & 359.38 & $8.33{\pm}0.05$ \\

8  & $359.76{\pm}8.92$ &&
$504^2$ & $359.76{\pm}8.92$ &&
S & 4,096 (1)     & 358.54 & $21.80{\pm}1.15$ \\

12 & $544.36{\pm}4.79$ &&
$518^2$ & $356.40{\pm}6.90$ &&
S & 8,192 (1)     & 361.27 & $36.85{\pm}0.32$ \\

16 & $765.93{\pm}6.41$ &&
& &&
S & 16,384 (1)    & 366.20 & $69.66{\pm}0.66$ \\

   & &&
& &&
S & 30,000 (1)    & 351.76 & $128.84{\pm}3.42$ \\

   & &&
& &&
S & 60,000 (2)    & 360.11 & $252.86{\pm}1.23$ \\

   & &&
& &&
S & 120,000 (4)   & 363.86 & $505.26{\pm}1.14$ \\

\addlinespace
   & &&
& &&
D-$252^2$ & 63,504 (3)  & 117.40 & $228.90{\pm}4.30$ \\

   & &&
& &&
D-$364^2$ & 132,496 (5) & 202.58 & $496.19{\pm}7.48$ \\

   & &&
& &&
D-$504^2$ & 254,016 (9) & 364.10 & $1029.28{\pm}5.10$ \\
\bottomrule
\end{tabular}
\caption{Measured runtime points underlying the scaling
results in the main paper. The clip-length sweep fixes
$H=W=504$ and $Q=4096$; the resolution sweep fixes
$T=8$ and $Q=4096$. The query sweep uses $T=8$;
S denotes sparse decoding at $504^2$, while D denotes
dense decoding at the indicated resolution. Parentheses
report the number of chunks with a maximum chunk size
of 30k. Results use FP16 on one NVIDIA A800 with five
warm-up runs, 20 measured repeats, and CUDA synchronization.}
\label{tab:runtime_raw}
\end{table*}
\noindent
At $504{\times}504$ with $Q{=}4096$, encoding scales
approximately linearly with clip length, with a fitted slope
of 45.70 ms/view ($R^2{=}0.995$). At fixed $T{=}8$,
encoding also scales with spatial resolution, with a slope
of 1.26 s per megapixel per clip ($R^2{=}0.995$).
Across the sparse query-count sweep, encoding remains nearly
constant (CV $= 1.20\%$), confirming that source--target
selection and query decoding occur after joint encoding.

Query-dependent decoding scales linearly with $Q$, with a
slope of 4.18 ms per 1k queries ($R^2{=}1.000$).
At $504{\times}504$, pyramid construction adds approximately
12.19 ms per chunk and is nearly constant across the sparse
sweep (CV $= 0.34\%$). Dense decoding follows the same
approximately linear trend ($R^2{=}0.999$). At $Q{=}60$k,
30k-query chunking reduces peak memory from 15.68 GB to
12.52 GB, while changing time per query by only $-1.87\%$.

\subsection{Cross-Attention Localization on Long Sequences}
\label{sec:supp-cross-attention}

\begin{figure*}[!t]
\centering
\includegraphics[width=0.95\textwidth]{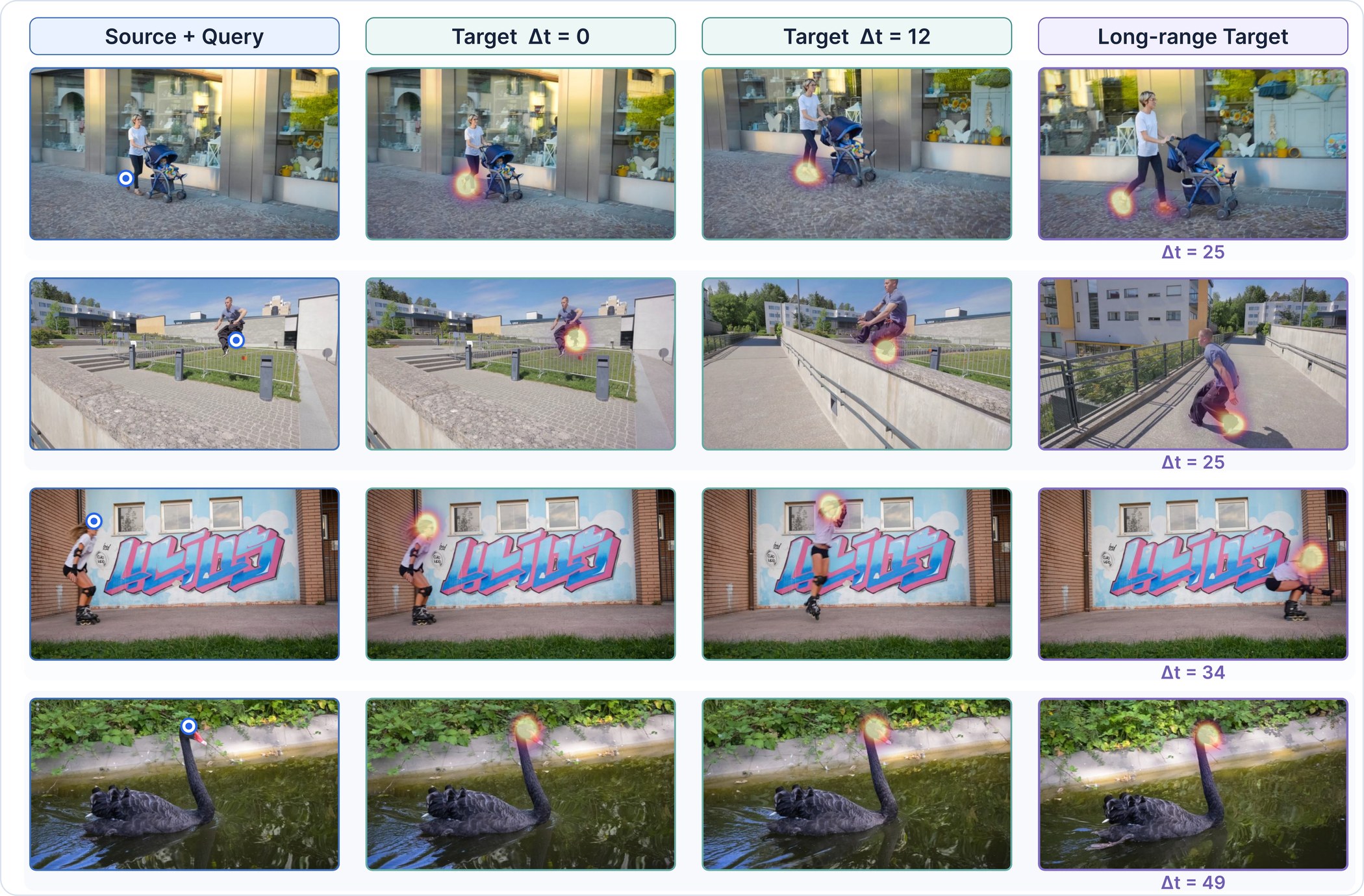}
\caption{First-layer source-to-target cross-attention over increasing
temporal gaps. Each row fixes a source query and visualizes its
normalized attention on progressively more distant target frames.}
\label{fig:supp-cross-attention}
\end{figure*}
\noindent
Figure~\ref{fig:supp-cross-attention} visualizes the first
source-to-target cross-attention layer of the query decoder. For each
row, we fix a continuous source query and plot its normalized attention
over the full target feature map as the temporal gap $(s,t)$ increases.
We use this visualization to examine whether pairwise cross-attention
recovers \emph{geometric} correspondences between frames rather than
merely aligning semantically similar regions.

Across the examples, the peak response consistently localizes to the
physically corresponding target location, including under partial
occlusion: when the matched point is not directly visible, attention
remains concentrated near the correct image region instead of drifting
to unrelated areas with similar appearance. This suggests that
source-to-target interaction carries useful 3D correspondence cues, not
just category-level semantic matching. The pattern also persists as the
source--target separation grows, indicating that our pairwise design
remains effective on longer sequences without a fixed-horizon temporal
embedding.

We also observe a recurring failure mode that highlights a current
limitation. When the query is placed on one symmetric body part, such
as the right foot, the map often assigns secondary mass to the
semantically and visually similar counterpart (e.g., the left foot).
Similar ambiguities appear for left/right hands. The model can therefore
confuse mirror-symmetric extremities over long horizons, and stable
long-term identity maintenance for such parts remains an open challenge.

\subsection{Backbone Initialization}
\label{sec:supp-backbone-init}

Table~\ref{tab:supp_backbone_init} compares the internal encoder of
Section~\ref{sec:supp-backbone} with publicly available
VGGT-$\Omega$~\cite{wang2026vggtomega} and
DA3-Giant~\cite{lin2025depthanything3} as alternative initializations.
All three variants use the \emph{same} UniQuery4R training data and
recipe, trained for $150$k iterations on $8{\times}$ NVIDIA A800 GPUs
(distinct from the final $16{\times}$ H20 checkpoint reported in the
main paper).
\begin{table}[!t]
\centering
\setlength{\tabcolsep}{4pt}
\begin{tabular}{@{}lcc@{}}
\toprule
Encoder initialization & APD$\uparrow$ & EPE$\downarrow$ \\
\midrule
VGGT-$\Omega$~\cite{wang2026vggtomega} & 80.88 & 0.1857 \\
DA3-Giant~\cite{lin2025depthanything3} & 81.29 & 0.1821 \\
Internal foundation model & \textbf{83.62} & \textbf{0.1597} \\
\bottomrule
\end{tabular}
\caption{WorldTrack dynamic-point macro-average under matched training
($150$k iterations, $8{\times}$ A800) with different encoder
initializations.}
\label{tab:supp_backbone_init}
\end{table}

\noindent
Even with fully \emph{public} initializations, UniQuery4R already
achieves strong WorldTrack dynamic-point macro-averages under this
matched setup: VGGT-$\Omega$ and DA3-Giant reach APD/EPE of
$80.88/0.186$ and $81.29/0.182$, respectively, outperforming the
other compared feed-forward methods in Table~\ref{tab:dynamic_points} of the main paper
(best baseline: V-DPM, $80.13/0.194$). This indicates
that much of the reported gain comes from the UniQuery4R query framework
rather than from the internal encoder alone. The internal foundation
model still yields the best numbers ($83.62/0.160$), providing an
additional but not strictly necessary improvement when public weights
are available.

\paragraph{SynthVerse Zero-Shot Evaluation.}
SynthVerse~\cite{zhao2026synthverse} is a recently released,
open-source synthetic dataset for 2D and 3D point tracking.
It contains approximately $48$K sequences and $5.816$M training
frames, covering articulated and deformable objects, humans,
animals, navigation, embodied manipulation, animated-film
content, and hand--object interaction. The dataset includes
both egocentric and allocentric views. Its official benchmark
contains seven domain subsets---Nav, Human, Animal, Objects,
Embodied, Film, and Interaction---and reports their aggregate
performance as mAverage.

To evaluate transfer under the same protocol used in our
WorldTrack experiments, we select $50$ SynthVerse-Benchmark
sequences with at least $64$ frames and evaluate the first $64$
frames of each sequence. We use frame~$0$ as the reference,
apply global median-norm scale alignment, and report EPE over
dynamic points, following the protocol in
Section~\ref{sec:supp-extra-results}. This evaluation differs
from the official SynthVerse benchmark, which reports
$\mathrm{AJ}_{2D}$, $\mathrm{APD}_{2D}$, $\mathrm{AJ}_{3D}$,
$\mathrm{APD}_{3D}$, and occlusion accuracy. Unless otherwise
specified, none of the models in Table~\ref{tab:supp_synthverse}
is trained or fine-tuned on SynthVerse.

\begin{table}[!t]
\centering
\setlength{\tabcolsep}{4pt}
\begin{tabular}{@{}lc@{}}
\toprule
Method & EPE$\downarrow$ \\
\midrule
SpaTrackerV2~\cite{xiao2025spatialtrackerv2} & 5.5187 \\
St4RTrack~\cite{feng2025st4rtrack} & 3.8485 \\
TraceAnything~\cite{liu2025traceanything} & 7.3385 \\
Any4D~\cite{karhade2025any4d} & 5.2362 \\
V-DPM~\cite{sucar2026vdpm} & \underline{2.7367} \\
4RC~\cite{luo20264rc} & 3.2957 \\
OpenD4RT~\cite{opend4rt} & 4.7392 \\
\midrule
\textbf{UniQuery4R (zero-shot)} & \textbf{2.4105} \\
\midrule
UniQuery4R (+ SynthVerse train) & 1.3682 \\
\bottomrule
\end{tabular}
\caption{Dynamic-point EPE (m) on SynthVerse ($50$ sequences $\times$ $64$
frames; WorldTrack protocol). Zero-shot evaluation excludes SynthVerse
training; the last row uses the official training split with benchmark
sequences held out.}
\label{tab:supp_synthverse}
\end{table}

\noindent
Without SynthVerse training, UniQuery4R obtains an EPE of
$2.41$\,m, compared with $2.74$\,m for the strongest baseline,
V-DPM, corresponding to a $12.0\%$ relative reduction. Adding
the official SynthVerse training split reduces the EPE of
UniQuery4R to $1.37$\,m, a further reduction of $43.2\%$ relative
to its zero-shot result. This improvement indicates a
substantial domain gap between our original training data and
SynthVerse, while also showing that SynthVerse training data
provides useful supervision for this evaluation domain.

\FloatBarrier
\section{Additional Qualitative Results}
\label{sec:supp-qualitative}

We supplement the main-paper DAVIS example
(Figure~\ref{fig:davis_comparison}) with additional tracking,
correspondence, motion masks, and static reconstruction.

\subsection{3D Dynamic Tracking}

\begin{figure*}[t]
\centering
\includegraphics[width=0.92\textwidth]{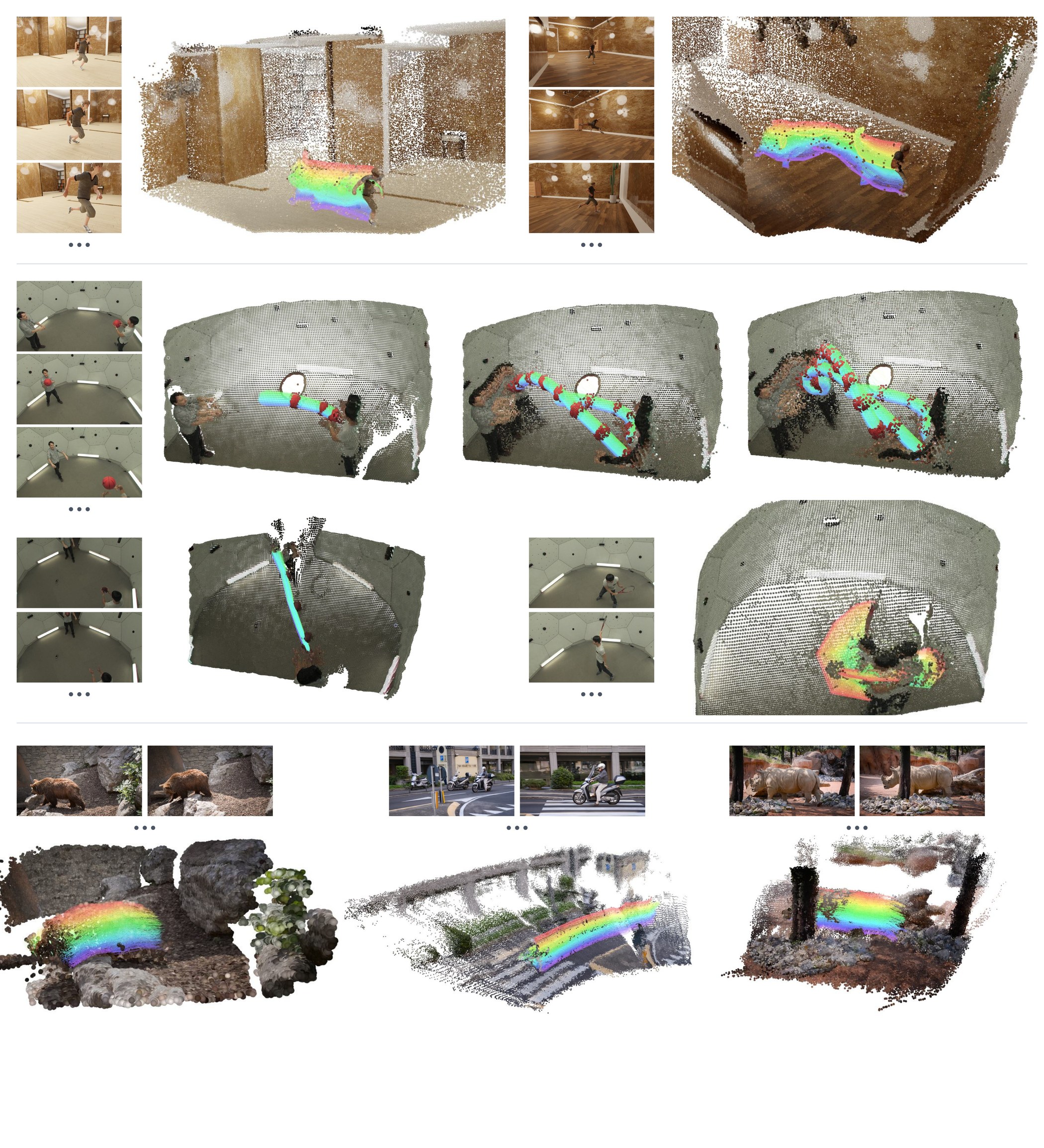}
\caption{Additional 3D tracking on diverse dynamic scenes. Each color
denotes one fixed source query over time.}
\label{fig:supp-more-qualitative}
\end{figure*}

Figure~\ref{fig:supp-more-qualitative} covers human motion, articulated
objects, and non-rigid deformation. Each color tracks one fixed source
query in the view-$0$ frame; trajectories remain coherent under occlusion
and viewpoint change.

\subsection{2D Correspondence}

\begin{figure*}[t]
\centering
\includegraphics[width=0.92\textwidth]{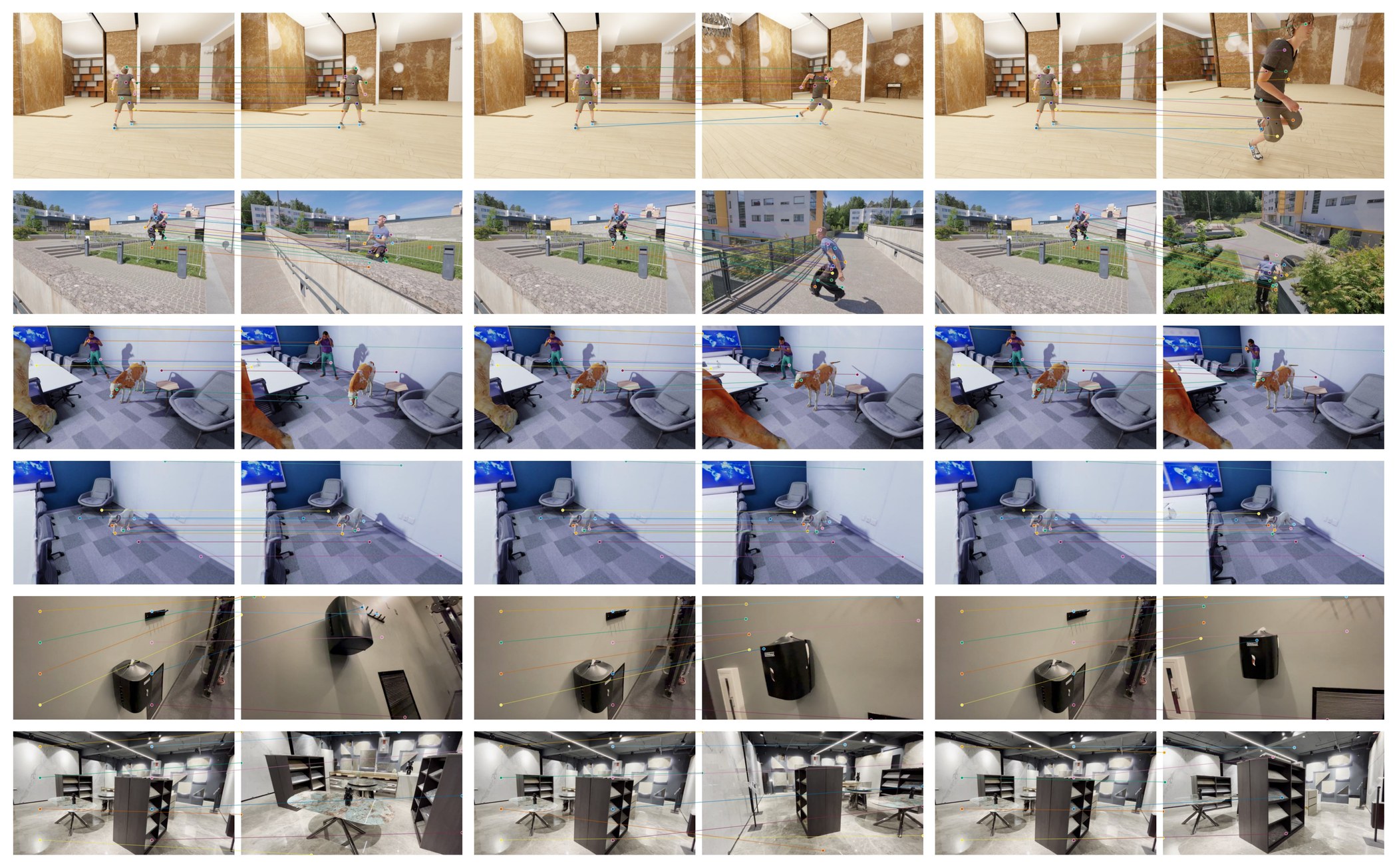}
\caption{Qualitative 2D correspondence for selected source--target
pairs.}
\label{fig:supp-2d-vis}
\end{figure*}

The query decoder predicts target-image coordinates for any
source--target pair (Figure~\ref{fig:supp-2d-vis}). Correspondences
follow object boundaries and stay on the target object under large motion
and partial occlusion.

\subsection{Motion Masks from Scene Flow}

\begin{figure*}[t]
\centering
\includegraphics[width=0.92\textwidth]{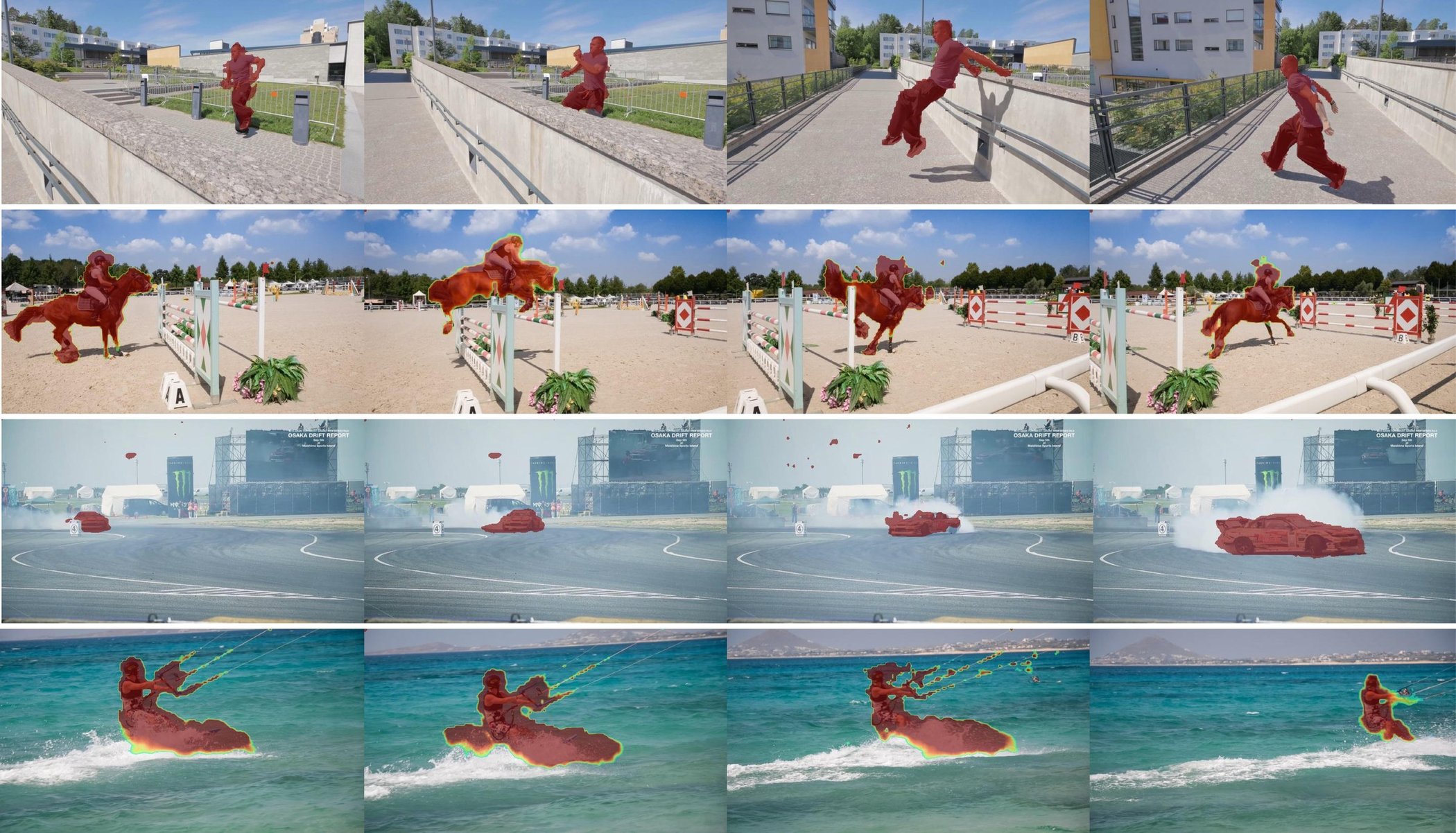}
\caption{Motion masks from thresholded consecutive scene flow
($\gamma{=}0.02$\,m).}
\label{fig:supp-motion-mask}
\end{figure*}

Motion masks are obtained without a dedicated segmentation head.
Given an $N$-frame clip (views $0,\ldots,N{-}1$), we encode all frames
once. After encoding, we construct dense grid queries on consecutive
source--target pairs
$(0,1),(1,2),\ldots,(N{-}2,N{-}1)$ and the single reverse pair
$(N{-}1,N{-}2)$. Decoding each pair yields a dense scene flow
$\Delta\mathbf{P}$ from the same direction--magnitude head used at
training time. A pixel is marked dynamic if
$\|\Delta\mathbf{P}\|$ exceeds $\gamma{=}0.02$\,m after metric-scale
restoration and static otherwise; short forward pairs
$(n,n{+}1)$ capture frame-to-frame motion, while the reverse pair
$(N{-}1,N{-}2)$ stabilizes the mask at the clip end.
Figure~\ref{fig:supp-motion-mask} shows that this simple threshold
recovers reasonable foreground motion regions despite sharing the same
encoder and decoder used for sparse queries.

\subsection{Static-Scene Reconstruction}

\begin{figure*}[t]
\centering
\includegraphics[width=0.88\textwidth]{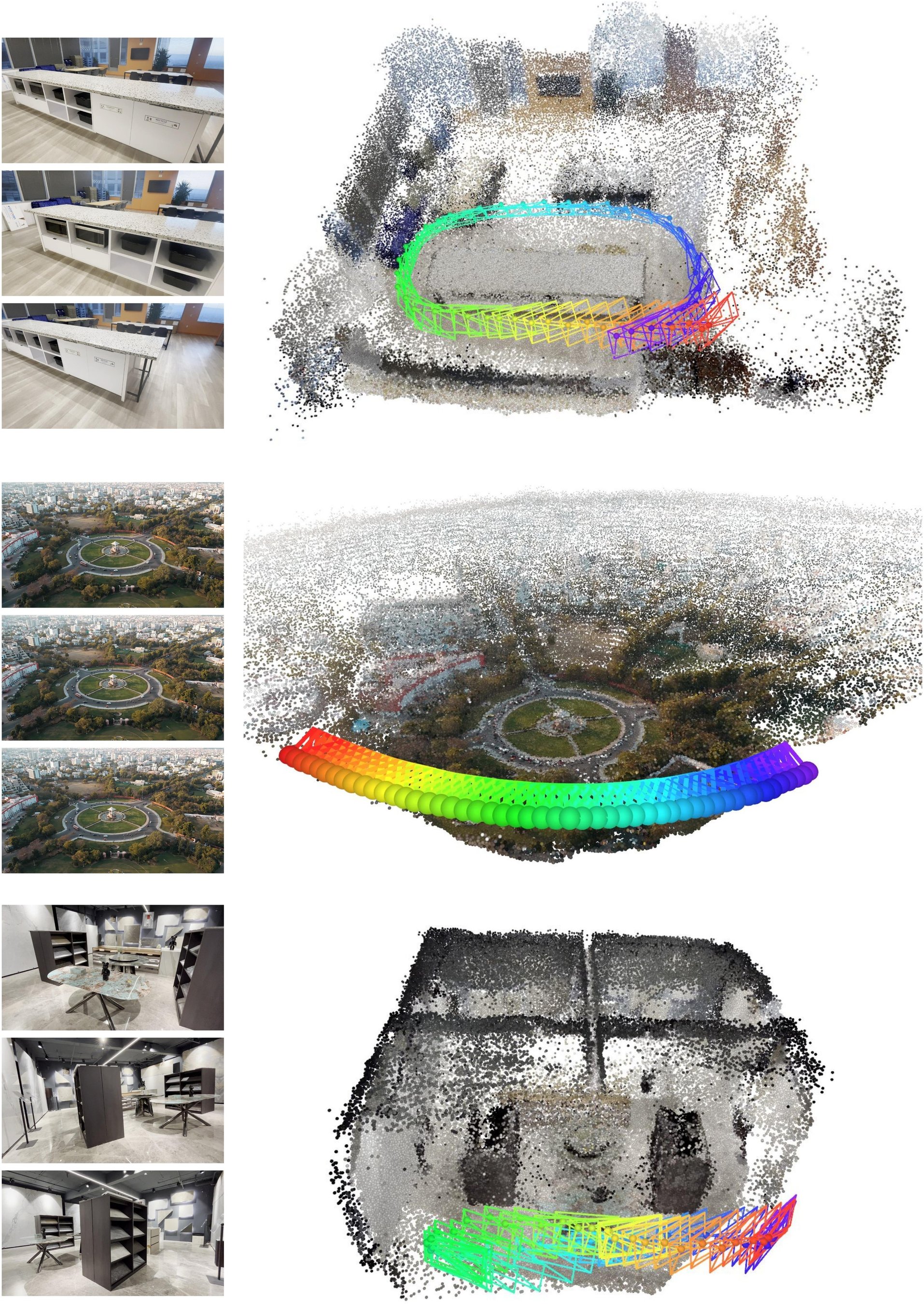}
\caption{Static-scene reconstruction with estimated cameras.}
\label{fig:supp-static-scenes}
\end{figure*}

The shared encoder and per-frame camera head also reconstruct static
scenes (Figure~\ref{fig:supp-static-scenes}): estimated poses and fused
point clouds on indoor and outdoor clips without task-specific finetuning.

\end{document}